%% file: main.tex
\documentclass[10pt,twocolumn,letterpaper]{article}

\usepackage{cvpr}      

\input{preamble}

\definecolor{cvprblue}{rgb}{0.21,0.49,0.74}
\usepackage[pagebackref,breaklinks,colorlinks,citecolor=cvprblue]{hyperref}

\usepackage{indentfirst}
\usepackage{multicol}
\usepackage{multirow}
\usepackage{hyperref}
\usepackage{times}  
\usepackage{xcolor}
\hypersetup{
    linkcolor=blue        
}

\def\paperID{10197} 
\def\confName{CVPR}
\def\confYear{2025}

\title{Real-IAD D³: A Real-World 2D/Pseudo-3D/3D Dataset for Industrial Anomaly Detection}

\author{  
    Wenbing Zhu$^{1,4}$\thanks{Equal contribution. $\dag$ Corresponding author.}, Lidong Wang$^{1*}$, Ziqing Zhou$^{1*}$, Chengjie Wang$^{2,3*}$,   
    Yurui Pan$^{1}$, Ruoyi Zhang$^4$,\\   
    Zhuhao Chen$^1$, Linjie Cheng$^1$, Bin-Bin Gao$^3$, Jiangning Zhang$^3$, Zhenye Gan$^3$, Yuxie Wang$^6$, \\
    Yulong Chen$^2$, Shuguang Qian$^4$, Mingmin Chi$^{1\dag}$,   
    Bo Peng$^{5\dag}$, Lizhuang Ma$^{2\dag}$  
    \\
    \small $^1$Fudan University \hspace{2em} $^2$Shanghai Jiao Tong University \hspace{2em} $^3$YouTu Lab, Tencent \\
    \small $^4$Rongcheer Co., Ltd. \hspace{2em} $^5$Shanghai Ocean University \hspace{2em} $^6$Suzhou University   
    \\
    \tt \small \{wbzhu23, ldwang23, zqzhou23, yrpan24, ljcheng24, zhuhaochen24\}@m.fudan.edu.cn, \\
    \tt \small\{jasoncjwang, danylgao, wingzygan, vtzhang\}@tencent.com, \{ruoyi.zhang, Bruce.qian\}@rongcheer.com, \\
    \tt \small 2309401037@stu.suda.edu.cn, \{llong\_c, lzma\}@sjtu.edu.cn, bpeng@shou.edu.cn,  mmchi@fudan.edu.cn  
}   
\vspace{-2em}
\begin{document}
\maketitle
\input{sec/0\_abstract}

\input{sec/1\_intro}

\input{sec/2\_related_work}

\input{sec/3\_method}

\input{sec/4\_experiment}
\input{sec/5\_conclusion}

{
    \small
    \bibliographystyle{ieeenat_fullname}
    \bibliography{main}
}
\input{sec/X\_suppl}


\end{document}

%% file: preamble.tex
\usepackage[dvipsnames]{xcolor}


%% file: sec/0_abstract.tex
\begin{abstract}

The increasing complexity of industrial anomaly detection (IAD) has positioned multimodal detection methods as a focal area of machine vision research. However, dedicated multimodal datasets specifically tailored for IAD remain limited. Pioneering datasets like MVTec 3D have laid essential groundwork in multimodal IAD by incorporating RGB+3D data, but still face challenges in bridging the gap with real industrial environments due to limitations in scale and resolution. To address these challenges, we introduce Real-IAD D³, a high-precision multimodal dataset that uniquely incorporates an additional pseudo-3D modality generated through photometric stereo, alongside high-resolution RGB images and micrometer-level 3D point clouds.
Real-IAD D³ features finer defects, diverse anomalies, and greater scale across 20 categories, providing a challenging benchmark for multimodal IAD
Additionally, we introduce an effective approach that integrates RGB, point cloud, and pseudo-3D depth information to leverage the complementary strengths of each modality, enhancing detection performance. Our experiments highlight the importance of these modalities in boosting detection robustness and overall IAD performance. 
The dataset and code are publicly accessible for research purposes at 
\href{https://realiad4ad.github.io/Real-IAD\_D3/}{https://realiad4ad.github.io/Real-IAD\_D3}

\end{abstract}

%% file: sec/1_intro.tex
\section{Introduction}

Anomaly detection is essential for ensuring product quality and reliability in industrial operations. Advances in computer vision and industrial AI have greatly enhanced the accuracy and efficiency of identifying and classifying anomalies. As industrial manufacturing underpins societal progress, rigorous quality control is essential~\cite{cao2024survey}. Defects arising during component production can substantially undermine product quality and lifespan, particularly in sensitive industries such as pharmaceuticals, food production, and battery manufacturing, where such defects present considerable safety risks to consumers~\cite{gudovskiy2022cflow,deng2022anomaly}.


\begin{figure}[t!]
  \centering
  \includegraphics[width=0.47\textwidth]{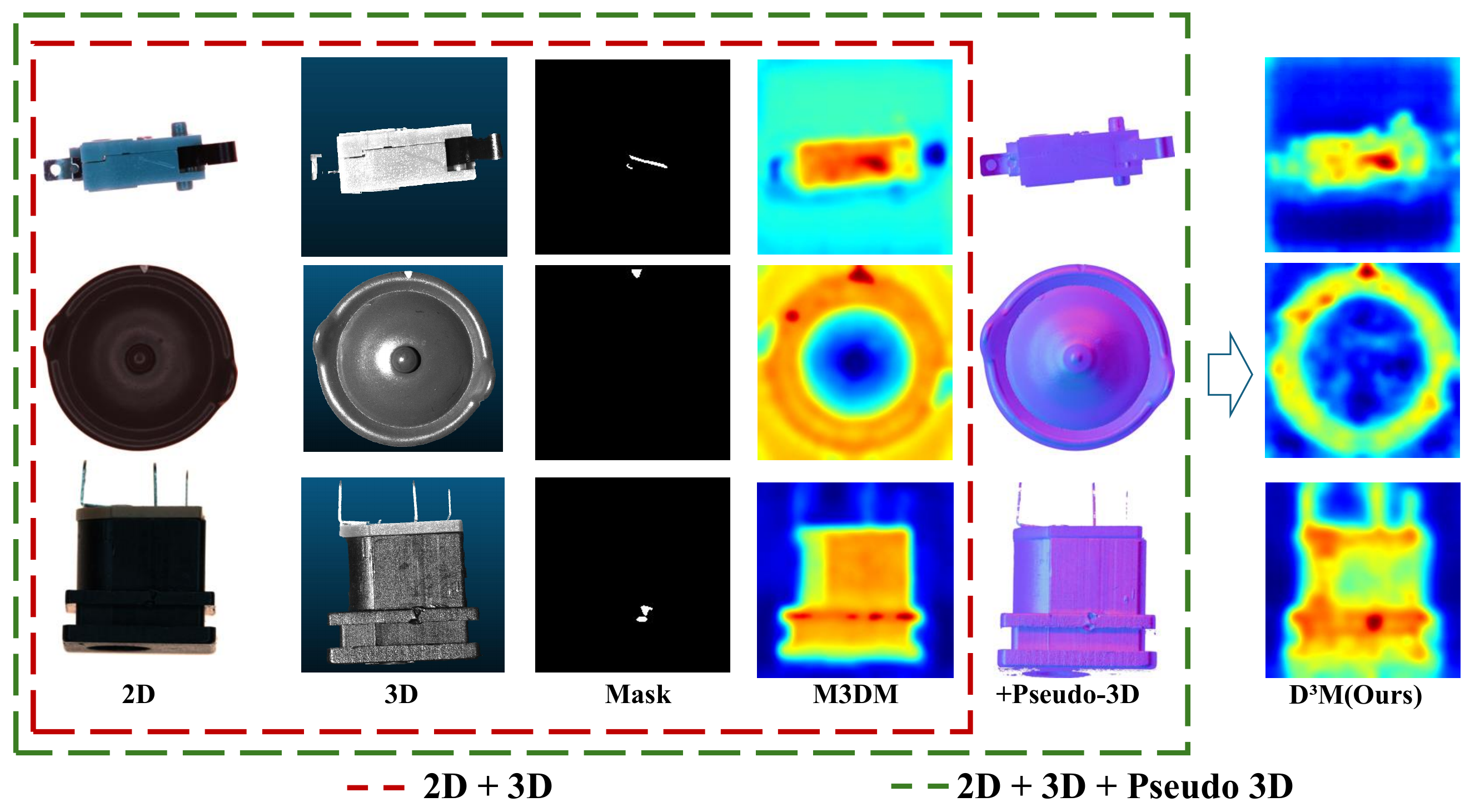}
  \caption{Pseudo-3D enhances defect localization over 2D and 3D.}
\label{introduction}
\vspace{-2em}
\end{figure}

To mitigate these risks, Industrial Anomaly Detection (IAD) has become an indispensable tool for maintaining product quality and operational safety across various industries. The development of datasets such as MVTec AD~\cite{bergmann2019mvtec}, VisA~\cite{zou2022spot} and Real-IAD~\cite{wang2024real} marked a significant milestone in anomaly detection, enabling unsupervised learning techniques that model normal sample distributions and classify outliers as anomalies. However, 2D image-based anomaly detection often falls short in industrial contexts, particularly for defects like scratches and dents, which are highly sensitive to variations in color and texture.


Recognizing these challenges, several multimodal datasets that integrate 2D and 3D point-cloud data have been developed to more effectively capture the complexity of real-world industrial environments. For instance, MVTec 3D-AD~\cite{bergmann2021mvtec} is designed for unsupervised 3D anomaly detection and localization, targeting geometric anomalies such as scratches, dents, and contaminations across 10 object categories. Another noteworthy dataset, Eyecandies~\cite{eyecandies}, introduces synthetic images of 10 candy-like objects with precise 2D, depth, and normal map annotations, offering automated and unbiased labeling using synthetic data.


With the expansion of multimodal datasets, new approaches in industrial anomaly detection have emerged, leveraging the combined strengths of 2D and 3D data to enhance detection accuracy and robustness in complex industrial environments. Examples include Shape-Guided Dual-Memory Learning, which detects subtle surface irregularities and volumetric anomalies by learning both 3D structures and 2D visual features~\cite{chu2023shape}, and symmetric student-teacher networks that use a teacher-student model to detect deviations from learned normal distributions, enabling more effective unsupervised detection~\cite{rudolph2023asymmetric}. Hybrid approaches further integrate 2D and point cloud data to capture both surface and structural anomalies, offering a more robust detection mechanism~\cite{wang2023multimodal}. Techniques like Total Recall in Industrial Anomaly Detection expand anomaly detection recall by storing normal patterns in memory networks, thereby increasing the range of normal data distributions~\cite{roth2022towards}, and Cheating Depth incorporates synthetic depth maps with 2D and point cloud data to enhance the detection of subtle surface anomalies~\cite{zavrtanik2024cheating}.


Despite these advances, existing datasets still exhibit significant limitations that hinder their practical applicability in industrial anomaly detection. Most notably, current datasets often have insufficient point-cloud resolution to capture fine details, and the limited diversity in material types and defect categories restricts models' generalizability to real-world applications. Consequently, the high anomaly detection metrics reported on these datasets may not accurately reflect performance in actual industrial environments, where conditions are much more variable.



To address these limitations, we introduce Real-IAD D³, a larger high-precision multimodal dataset that uniquely incorporates an additional pseudo-3D modality generated through photometric stereo, alongside high-resolution RGB images and micrometer-level 3D point clouds. Real-IAD D³ includes 20 object categories, each containing 4–6 distinct defect types, and provides point-cloud data with a resolution as fine as 0.002 mm across 8,450 samples. Specifically, it comprises 5,000 normal samples and 3,450 anomalous samples, with each sample containing synchronized RGB, point cloud, and pseudo-3D images. The anomalies are designed to closely simulate real industrial scenarios, making them more challenging and practical for both single-modal and multi-modal applications. This substantial increase in dataset scale and diversity introduces greater challenges for anomaly detection algorithms, fostering improved robustness and accuracy.


A distinctive feature of Real-IAD D³ is its photometric stereo-derived pseudo-3D data, addressing limitations in existing datasets by capturing subtle defect characteristics influenced by material properties. Extensive experiments demonstrate that D³ multimodal data significantly elevates IAD detection standards, offering a comprehensive, challenging benchmark for multimodal anomaly detection. Additionally, we propose a benchmark model for D³ anomaly detection, establishing a foundational reference for future research in this field.


Overall, our contributions are summarized as follows: 
\begin{itemize} 
\item We introduce Real-IAD D³, a high-precision, strictly-aligned multimodal dataset that synchronizes RGB and 3D point cloud data with carefully curated, challenging defects. Comprising 8,450 samples across 20 diverse industrial categories, this dataset establishes a new benchmark in both scale and variety for multimodal industrial anomaly detection.
\item We introduce pseudo-3D modality from photometric stereo in Real-IAD D³, which offers relative depth information to improve pixel-level defect detection, particularly for subtle surface details like fine scratches and minor dents.
\item We introduce the D³M benchmark, which aligns 2D, pseudo-3D, and 3D modalities to provide a comprehensive representation of industrial components. This benchmark demonstrates the advantages of multimodal fusion for enhanced precision and reliability in IAD.
\end{itemize}

%% file: sec/2_related_work.tex
\section{Related Work}

\subsection{Multi-modal 3D-AD Methods}
Recent advances in 2D AD have introduced sophisticated methods~\cite{cao2024survey, rudolph2023asymmetric, roth2022towards, gudovskiy2022cflow, lei2023pyramidflow, defard2021padim, rippel2021modeling, bergmann2020uninformed, salehi2021multiresolution, deng2022anomaly, tien2023revisiting,wang2024pspu,wang2025softpatch+}, including image reconstruction, feature distillation, and few-shot learning. Building on this progress, MVTec 3D-AD~\cite{bergmann2019mvtec} has spurred interest in 3D anomaly detection (3D-AD), which, while promising, is still less developed~\cite{bergmann2019mvtec, zou2022spot, wang2023multimodal, chen2023easynet}. 
In 3D-AD, depth data is widely used to reduce background noise and complement RGB data. Bergmann et al.~\cite{bergmann2023anomaly} employ a teacher-student model, while Horwitz et al.~\cite{horwitz2023back} integrate 3D descriptors in a KNN framework. AST~\cite{rudolph2023asymmetric} combines depth-based background removal with 2D techniques, and Cheating Depth~\cite{zavrtanik2024cheating} simulates depth features to overcome RGB limitations. Shape-Guided Dual-Memory Learning~\cite{chu2023shape} incorporates shape information for better localization, and PatchCore~\cite{patchcore} excels in multimodal AD via feature matching. M3DM~\cite{wang2023multimodal} highlights RGB and point cloud fusion, showcasing multimodal methods' effectiveness in enhancing detection in complex settings.

\begin{figure*}[ht]
  \centering
  \includegraphics[width=0.95\textwidth]{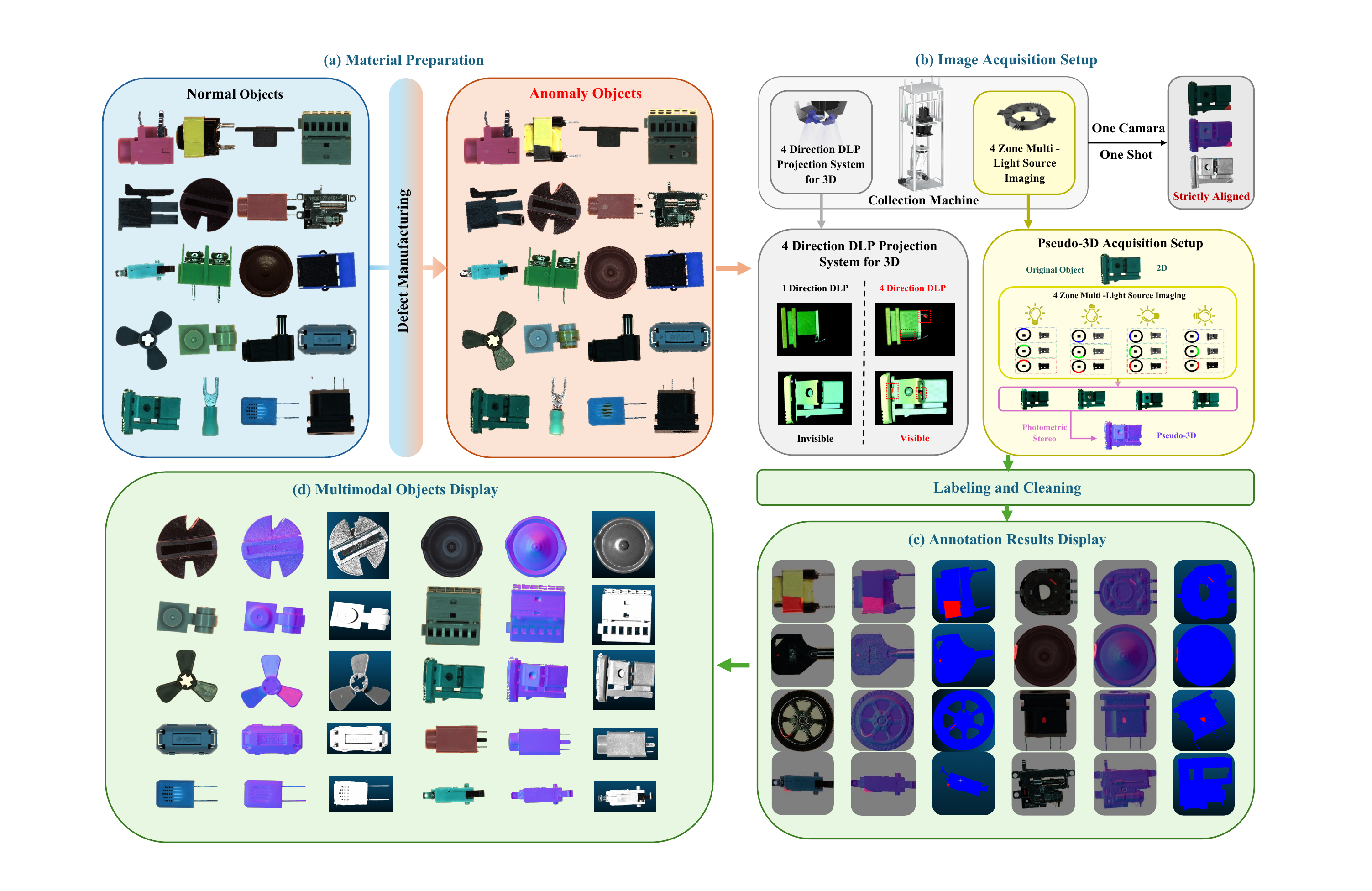}
  \caption{(a) Material preparation and defect creation across 20 product categories; (b) Image acquisition setup using a single camera, with a 4-zone light source to capture pseudo-3D images (via photometric stereo) and 4-direction DLP structured light to capture 3D images; (c) Data collection, annotation, and cleaning with pixel-level accuracy; (d) Multi-modal imaging demonstration, including 2D, pseudo-3D, and 3D data.}
  \label{fig1}
  \vspace{-1em}
\end{figure*}

\subsection{3D-AD Datasets.} Since 2007, the field of 2D anomaly detection (2D-AD) has seen substantial advancement, driven by numerous datasets that facilitate various methodologies, including image reconstruction~\cite{zavrtanik2024cheating, bergmann2021mvtec, rlr, uniad, diad}, feature distillation~\cite{cao2024survey, gudovskiy2022cflow, wan2022position}, and few-shot anomaly detection~\cite{xie2023graphcore, zhang2023augmentation, chen2023zero, winclip, PromptAD}. In contrast, 3D anomaly detection (3D-AD) remains a relatively nascent field, underpinned by only a limited number of foundational datasets. Notably, MVTec 3D-AD~\cite{bergmann2021mvtec} provides a benchmark dataset for unsupervised 3D anomaly detection with high-resolution depth scans and precise defect annotations across 10 object categories. Similarly, Eyecandies~\cite{eyecandies} offers a synthetic, photo-realistic dataset that includes RGB, depth, and normal maps, allowing for automated and unbiased defect labeling in controlled lighting conditions. Real3D-AD~\cite{liu2024real3d} further contributes a large-scale, high-precision dataset, specifically designed for industrial anomaly detection using point cloud data. Despite these contributions, existing 3D datasets are limited in their representation of industrial materials and defect types, underscoring the need for more diverse and comprehensive datasets to advance 3D-AD research.


%% file: sec/3_method.tex
\section{Real-IAD D³ Dataset Description}
The Real-IAD D³ dataset is an extensive multimodal industrial anomaly detection dataset, encompassing 20 distinct product categories and 69 defect groups. Each group contains an average of 50 samples, resulting in 3.45 defects per material. 
\begin{table*}[htbp]
  \centering
  \caption{Comparison of the proposed Real-IAD D³ dataset with existing 3D+2D(RGB) datasets, including MVTec-3D AD and Real3D-AD, across various parameters. }
  \resizebox{0.99\textwidth}{!}{
    \begin{tabular}{cccccccccc}
    \toprule
    Dataset & Product Categories & Defect Categories & Sample Number & 3D Point-Cloud Resolution & Point Precision & 3D Format & Photometric stereo & Multi-modal Sync & Multi-Direction DLP \\
    \midrule
    MVTec 3D-AD      & 10    & 33     & 4147   & 0.37mm & 0.11mm          & TIFF  & $\times$    & $\times$    & $\times$ \\
    Real3D-AD          & 12    & 40    & 1254  & 0.04mm & 0.011mm-0.015mm & ASC, PLY, STL, OBJ, IGES & $\times$    & $\times$    & $\times$ \\
    Real-IAD D³ (Ours) & 20    & 69    & 8450 & 0.01mm & 0.002mm         & ASC, PLY, STL, OBJ, IGES, TIFF & $\checkmark$   & $\checkmark$   & $\checkmark$ \\
    \bottomrule
    \end{tabular}%
  \label{tab1}}%
\end{table*}%

\begin{figure*}[h]
  \centering
  \includegraphics[width=\textwidth]{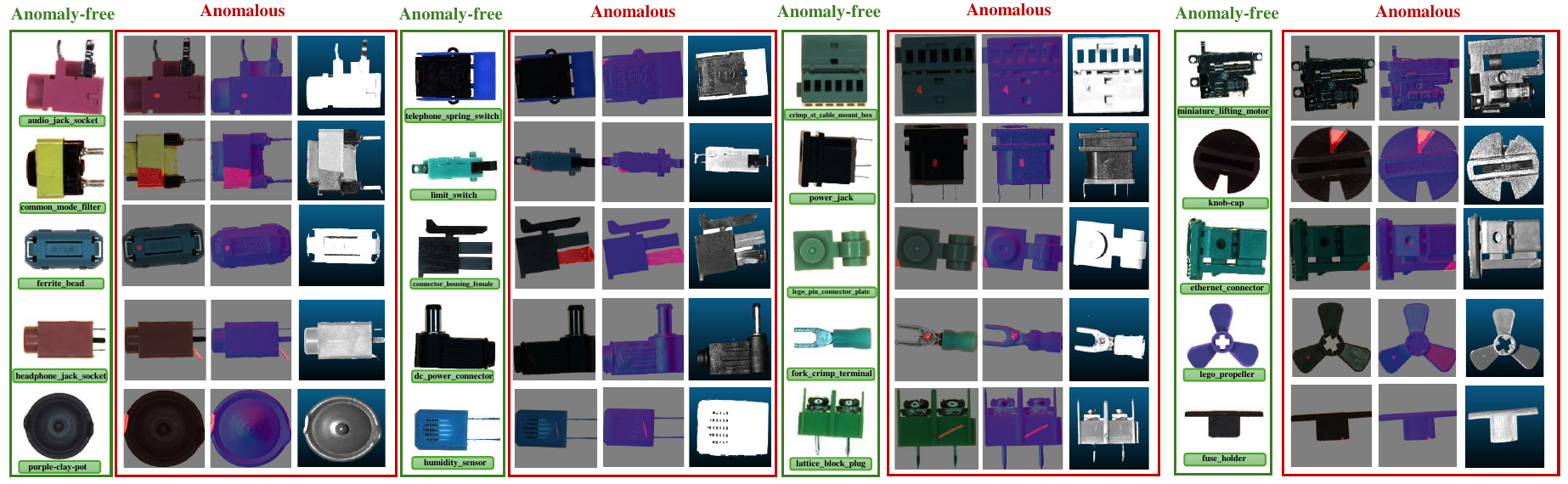}
  \caption{Examples of the 20 materials in the Real-IAD D³ dataset.  Each group of images represents a specific material, with the first column showing anomaly-free images.  The subsequent three columns display anomalous samples for each material in 2D (RGB images), pseudo-3D (pseudo-3D depth from photometric stereo), and 3D point cloud formats. 
}
\label{fig2}
\end{figure*}
In total, the dataset includes 8,450 samples: 5,000 normal samples and 3,450 abnormal samples. Each sample includes synchronized 2D, pseudo-3D photometric stereo fusion, TIFF, and PLY data.
The dataset captures a wide range of defect area proportions, spanning from 0.46\% to 6.39\%. The multi-light-source setup supports four-point cloud resolutions, reaching up to 16.2 million points (5328x3040) with an accuracy of 0.002 mm. The proportion of defective points varies from 0.33\% to 7.34\%.

\subsection{Data Collection and Construction manner}

\textbf{Data Collection and Annotation.} We have curated a dataset comprising 20 industrial products across various material types, including metal, plastic, ceramics, and composites. These objects were carefully selected to cover a broad range of industrial scenarios.

For each object, we manually introduced several types of defects, such as scratches, dents, cracks, missing parts, and deformation. This variety ensures a realistic and challenging environment for anomaly detection. Both normal and defective samples were then prepared for multimodal data acquisition, ensuring a balanced representation of different defect types.

\textbf{Prototype Construction.} The acquisition setup consists of an integrated system capable of capturing synchronized 2D, pseudo-3D surface normals, and 3D point cloud data. As shown in Fig~\ref{fig1}, the acquisition system is designed to capture 2D, pseudo-3D, and 3D data using a unified setup, ensuring precise alignment and seamless interaction across modalities. A high-resolution camera with a resolution of 3,648 × 5,472 pixels is employed to capture detailed RGB images. For 3D data acquisition, as shown in Fig~\ref{fig1}(c) ,a four-direction structured light system is utilized to obtain highly accurate 3D point clouds, enabling the detection of fine surface details. A photometric stereo technique is also applied to generate pseudo-3D depth information by synthesizing surface normals from four directional light sources. This integrated system allows for comprehensive multimodal data capture, enhancing the precision and reliability of defect detection.

The photometric stereo process is based on capturing images of the object under different lighting conditions to compute the surface normals. As shown in Fig~\ref{fig1}(b), four distinct lighting directions are used, each producing an image with different shading effects. The intensity values from these images are then used to calculate the surface normals (\( \mathbf{n}(x, y) \)) at each pixel through the following photometric stereo equation:
\[
\mathbf{I}(x, y) = \mathbf{L} \cdot \mathbf{n}(x, y),
\]
\( \mathbf{I}(x, y) \) is the vector of intensity values for a given pixel \((x, y)\) from the four images. \( \mathbf{L} \) is the matrix of lighting directions, with each row representing the direction of one of the four light sources. \( \mathbf{n}(x, y) \) is the surface normal vector at the pixel location \((x, y)\).
Given the intensity values \( \mathbf{I}(x, y) \) and the known lighting directions \( \mathbf{L} \), the surface normal \( \mathbf{n}(x, y) \) can be computed by solving:
\[
\mathbf{n}(x, y) = (\mathbf{L}^\top \mathbf{L})^{-1} \mathbf{L}^\top \mathbf{I}(x, y).
\]



\begin{figure*}[h]
  \centering
  \includegraphics[width=0.8\textwidth]{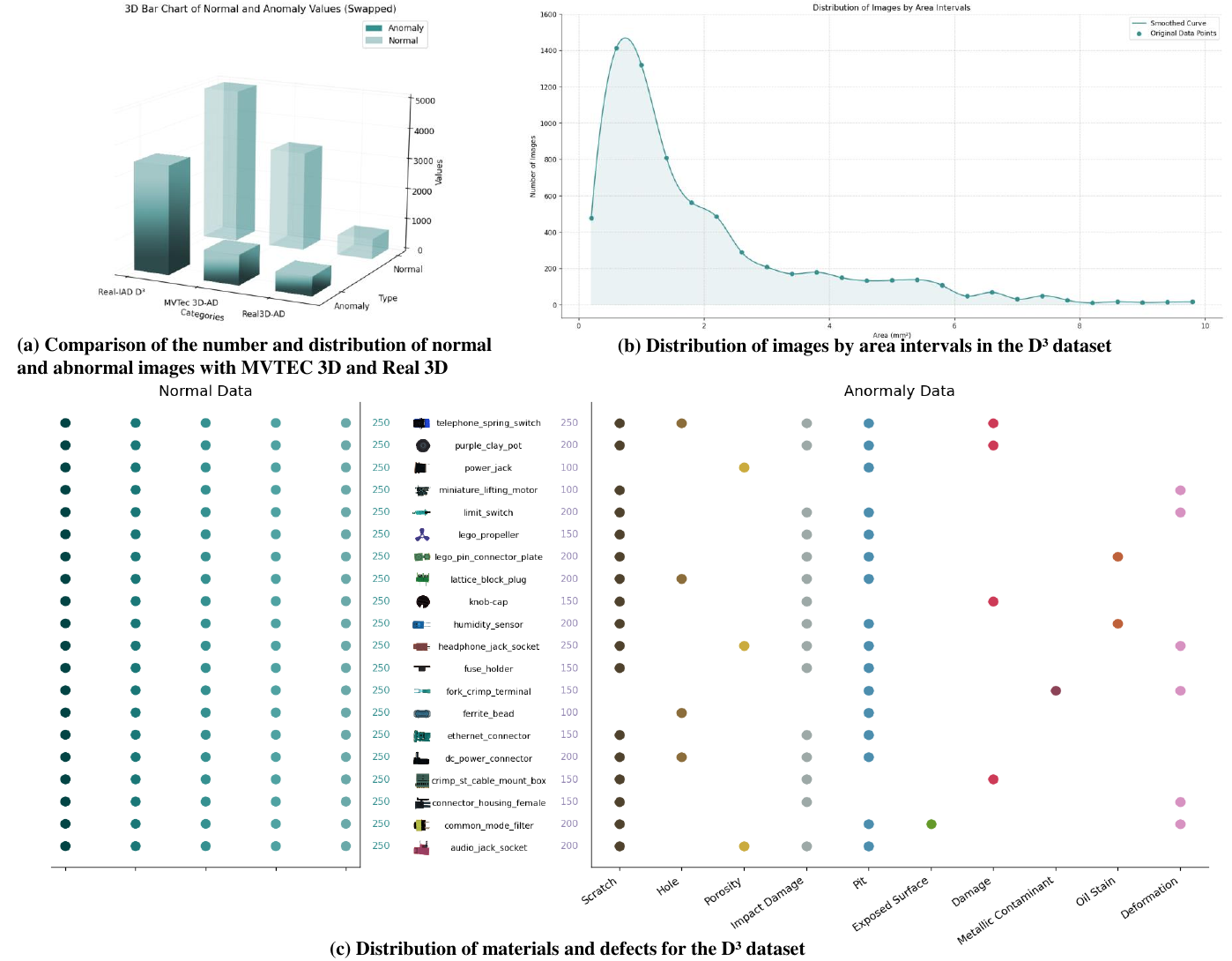}
  \caption{Statistical overview of the Real-IAD D³ dataset in comparison to MVTec 3D-AD and Real3D-AD, illustrating sample counts, normal/anomalous distribution per product, defect area ratios, and defect distribution patterns.}
\label{fig3}
\end{figure*}

\subsection{Comparison with Popular 3D Datasets}

Table \ref{tab1} provides a comparative overview of the Real-IAD D³ dataset alongside two benchmark datasets, MVTec 3D-AD and Real3D-AD, across several critical parameters. Real-IAD D³ offers extensive coverage with 20 product categories and 69 defect types, totaling 8,450 samples, which is substantially larger than MVTec 3D-AD (10 product categories, 33 defects) and Real3D-AD (12 product categories, 40 defects).
A distinct advantage of Real-IAD D³ is its fine point precision at 0.002 mm, outperforming the resolution and precision of MVTec 3D-AD (0.11 mm) and Real3D-AD (0.011 mm–0.015 mm). Additionally, Real-IAD D³ supports a variety of 3D formats (ASC, PLY, STL, OBJ, IGES, TIFF) and includes photometric stereo to enhance depth and capture fine surface details. Unlike MVTec 3D-AD and Real3D-AD, Real-IAD D³ captures RGB and 3D data on the same platform, ensuring natural alignment without additional calibration, and supports multi-directional DLP projection, making it a valuable dataset for industrial anomaly detection.

\subsection{Real-IAD D³ visualization}
The Real-IAD D³ dataset comprises 20 distinct product categories, covering a variety of industrial components, including mechanical parts, electronic devices, connectors, and sensors. This diversity enables a comprehensive evaluation of anomaly detection methods across different defect types. Each category includes three modalities: RGB images, 3D point clouds, and pseudo-3D surface normals from photometric stereo, providing a complete representation of the objects.As shown in Figure \ref{fig2}, this multi-modal approach—which includes original images, masked 2D images, pseudo-3D images with masks, and 3D point clouds—demonstrates the necessity of combining pseudo-3D and 3D modalities for thorough defect detection. On dark backgrounds, subtle defects like scratches and dents are challenging to identify in 2D images due to limited depth and detail. However, pseudo-3D images reveal these surface defects (e.g., on 'dc power connector' and 'power jack'), while 3D point clouds effectively capture larger geometric anomalies, such as those in 'lego propeller' and 'fuse holder.' This illustrates the complementary strengths of pseudo-3D and 3D data, ensuring accurate identification of both subtle and complex defects.


%% file: sec/4_experiment.tex
\section{Benchmark for D³ AD}

To address the challenges posed by both 2D and 3D anomaly detection, we propose a novel benchmark and anomaly detection framework called D³-Memory (D³M), which integrates 2D(RGB), point cloud, and pseudo-3D photometric stereo data. This system effectively leverages the strengths of each modality to enhance the detection of surface and structural anomalies, particularly those that are difficult to capture with conventional 2D or 3D data alone.  
\begin{figure}[h]  
  \centering  
  \includegraphics[width=0.48\textwidth]{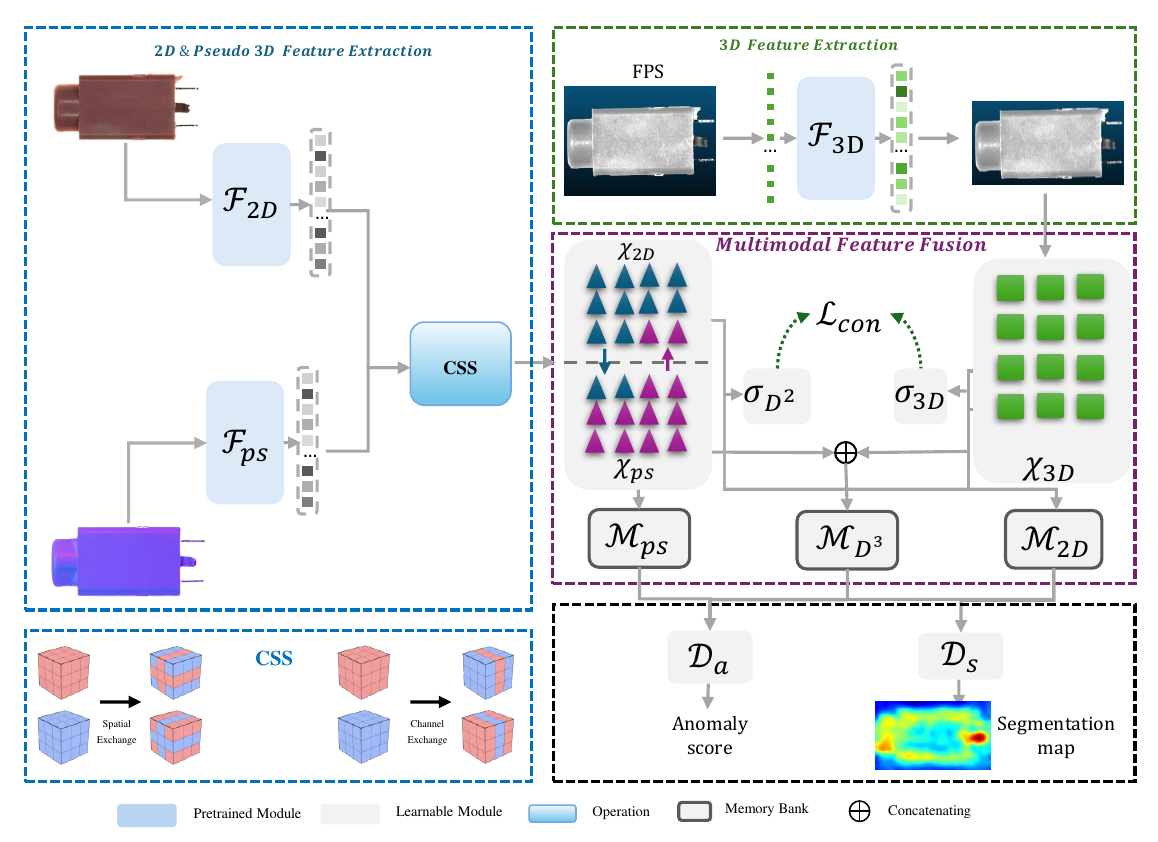}  
\caption{The D³M framework architecture for IAD, illustrating multi-modal feature extraction, feature fusion, and decision layer.}
\vspace{-1em}
\end{figure}  

\textbf{Multi-modal Feature Extraction.}
We use DINO (ViT-b/8)~\cite{wang2023multimodal} to extract essential visual features (texture, color, edges, surface normals) and PointMAE to capture geometric and depth features from 3D point cloud data.

To enhance the integration of 2D and photometric stereo data, we introduce Channel-Spatial Swapping(CSS) in our feature fusion strategy. This module swaps 10\% of the channel and spatial information between the RGB (\( X_{2D} \)) and photometric stereo (\( X_{PS} \)) feature maps, allowing for deeper interaction between visual and geometric features. Given feature maps \( X_{2D} \) and \( X_{PS} \), each of shape \( \mathbb{R}^{C \times H \times W} \), the swapping is defined as:

\[
X_{2D}^{swap} = (1 - \alpha) \times X_{2D} + \alpha \times X_{PS}^{c \leftrightarrow s}
\]
\[
X_{PS}^{swap} = (1 - \alpha) \times X_{PS} + \alpha \times X_{2D}^{c \leftrightarrow s}
\]
\[
X^s = \text{Block}_{k \times k}(X) \circ \alpha_{swap} \subseteq X^{c \leftrightarrow s}
\]

where \(\alpha = 0.1\) is the swapping ratio, $\circ$ indicates the composition operation, \( X^{c \leftrightarrow s} \) denotes a channel-spatial exchange, and  \( X^{s} \) is contained in \( X^{c \leftrightarrow s} \) describing a kernel-based spatial exchange, where the input is divided into $k \times k$ blocks and features are swapped with ratio $\alpha$ at the granularity of $\frac{H}{k} \times \frac{W}{k}$. This yields enriched features \( X_{2D}^{swap} \) and \( X_{PS}^{swap} \) that incorporate 10\% of each other’s properties, creating a more robust pseudo-3D representation for anomaly detection.

\textbf{Unsupervised Contrastive Feature Fusion.}  
To exploit the complementary nature of pseudo-3D and 3D modalities, we apply a fusion approach inspired by M3DM~\cite{wang2023multimodal}. Using an unsupervised contrastive learning method, we align features between 2D (with swapped pseudo-3D information) and 3D data to learn shared representations while preserving modality-specific details. Given a sample \( i \) and patches \( j \) from the swapped 2D feature map \( X_{2D}^{(i,j)} \) and the 3D feature map \( X_{3D}^{(i,j)} \), we apply a contrastive loss as follows:

\[
L_{con} = \frac{ \sum_{j=1}^{N_p} h_{2D}^{(i,j)} \cdot h_{3D}^{(i,j)} }{ \sum_{k=1}^{N_b} \sum_{j=1}^{N_p} h_{2D}^{(k,j)} \cdot h_{3D}^{(k,j)} }
\]

where \( h_{2D}^{(i,j)} \) and \( h_{3D}^{(i,j)} \) are MLP projections of the 2D and 3D feature vectors in patch \( j \) for sample \( i \). This fusion, resulting in the D³ representation \( M_{D³} \), promotes  between corresponding patches across modalities, capturing the strengths of each.

\textbf{Decision Layer Fusion.} Following the approach in M3DM~\cite{wang2023multimodal}, we fuse the 2D (swapped), pseudo-3D (swapped), and 3D features to create three memory banks: \( M_{2D} \), \( M_{PS} \), and fused \( M_{D³} \). Here, \( M_{2D} \) and \( M_{PS} \) store features from the channel-spatial swapped 2D and pseudo-3D modalities, while \( M_{D³} \) holds the fused representation learned through contrastive  of the swapped 2D (with pseudo-3D) and 3D features.

The final anomaly score \( a \) and segmentation map \( S \) are generated by comparing the features of a test sample with normal features in these memory banks using one-class SVM classifiers:

\[
a = D_a(\phi(M_{2D}, f_{2D}), \phi(M_{PS}, f_{PS}), \psi(M_{D^{3}}, f_{D^{3}}))
\]
\[
S = D_s(\psi(M_{2D}, f_{2D}), \psi(M_{PS}, f_{PS}), \psi(M_{D^{3}}, f_{D^{3}}))
\]

where \( \phi(\cdot) \) and \( \psi(\cdot) \) represent the anomaly scoring and segmentation functions, respectively.

\section{Experiment}

\subsection{Anomaly Detection on Real-IAD D³}

\textbf{ D³ $>$ 2D+3D.}
The experimental results in Table~\ref{tab2} validate the effectiveness of our D³ multimodal anomaly detection approach on the Real-IAD dataset. By integrating 2D (RGB), pseudo-3D, and 3D modalities, our method consistently surpasses single-modality (2D or 3D) and dual-modality (2D+3D) approaches, underscoring the value of additional modality information.
\begin{figure}[h]
  \centering
  \includegraphics[width=0.48\textwidth]{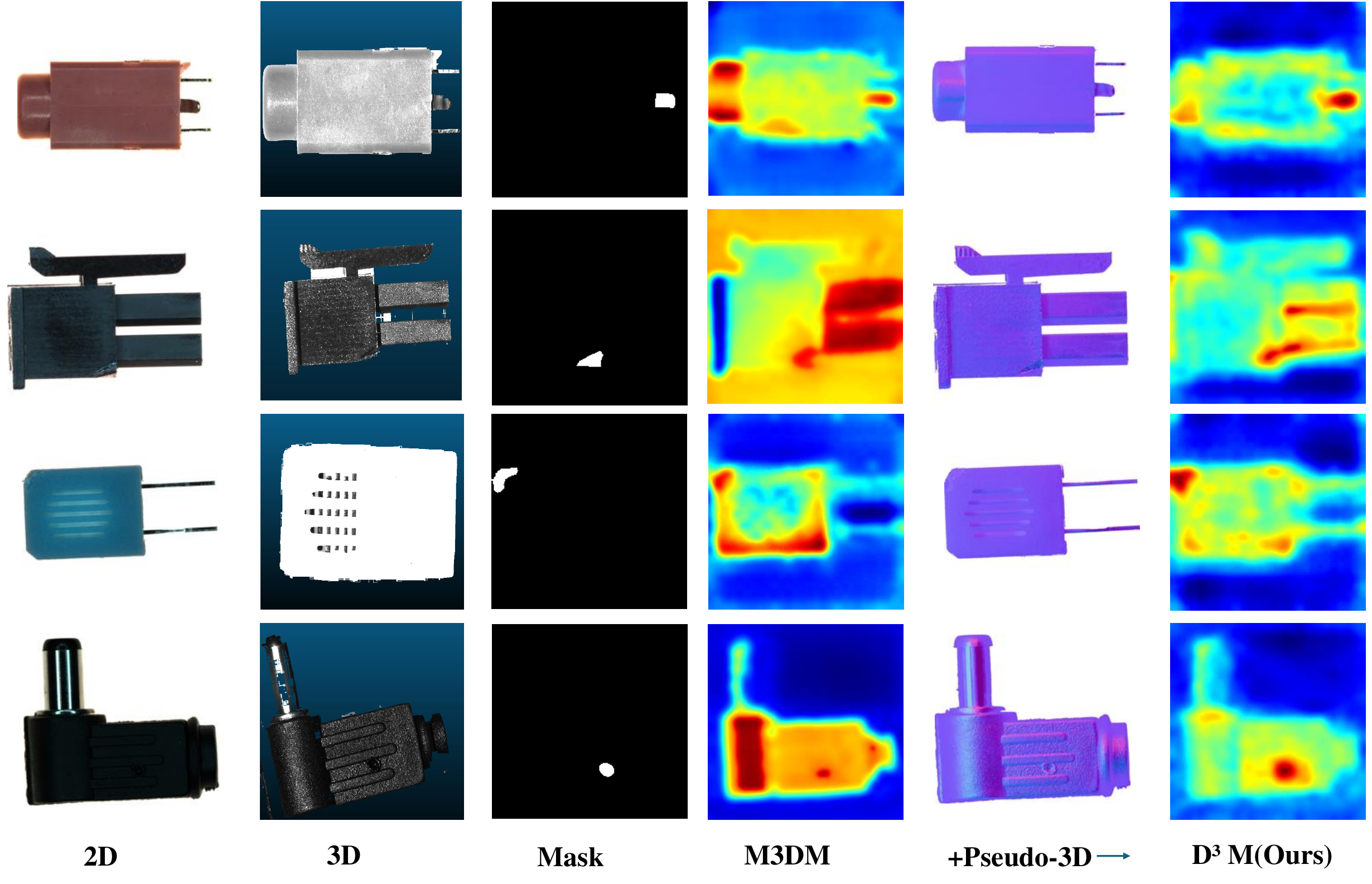}
\caption{Visualization of the effect of adding pseudo-3D to 2D+3D, showing enhanced segmentation maps.}
\label{vis1}
\vspace{-1em}

\end{figure}

\begin{table*}[htbp]
  \centering
  \caption{Performance of different multimodal anomaly detection methods on the Real-IAD dataset. The table presents the results in three different settings: Single Modality(RGB or Point-Cloud), 2D+3D, and D³. The two evaluation metrics are I-AUROC (image-level) and P-AUROC (pixel-level), with higher values indicating better performance. Our method consistently shows the best performance, especially in the D³ setting.}
    \resizebox{0.9\textwidth}{!}{
    \begin{tabular}{c|cccc|cc|cccccc|cc}
        \midrule
    Modality & \multicolumn{4}{c|}{RGB}      & \multicolumn{2}{c|}{3D} & \multicolumn{6}{c|}{2D+3D}                    & \multicolumn{2}{c}{D³} \\
    \midrule
    Model & \multicolumn{2}{c}{Cflow} & \multicolumn{2}{c|}{SimpleNet} & \multicolumn{2}{c|}{PointMAE} & \multicolumn{2}{c}{AST} & \multicolumn{2}{c}{PointMAE+PatchCore} & \multicolumn{2}{c|}{M3DM} & \multicolumn{2}{c}{D³M (Ours)} \\
    \midrule
    Metrics & I-AUROC & P-AUROC & I-AUROC & P-AUROC & I-AUROC & P-AUROC & I-AUROC & P-AUROC & I-AUROC & P-AUROC & I-AUROC & P-AUROC & I-AUROC & P-AUROC \\
    \midrule
    audio\_jack\_socket          & 0.943 & 0.944 & 0.973 & 0.926 & 0.763 & 0.655 & 0.860 & 0.590 & 0.926 & 0.673 & 0.981 & 0.699 & 0.983 & 0.757 \\
    common\_mode\_filter         & 0.271 & 0.847 & 0.717 & 0.822 & 0.725 & 0.687 & 0.899 & 0.802 & 0.523 & 0.922 & 0.580 & 0.934 & 0.618 & 0.947 \\
    connector\_housing-female    & 0.839 & 0.921 & 0.795 & 0.891 & 0.958 & 0.428 & 0.914 & 0.716 & 0.870 & 0.919 & 0.920 & 0.979 & 0.931 & 0.951 \\
    crimp\_st\_cable\_mount\_box & 0.18  & 0.442 & 0.372 & 0.745 & 0.291 & 0.363 & 0.485 & 0.589 & 0.713 & 0.931 & 0.749 & 0.933 & 0.811 & 0.969 \\
    dc\_power\_connector         & 0.661 & 0.726 & 0.661 & 0.725 & 0.849 & 0.507 & 0.995 & 0.770 & 0.720 & 0.921 & 0.715 & 0.950 & 0.922 & 0.947 \\
    ethernet\_connector          & 0.967 & 0.853 & 0.981 & 0.866 & 1     & 0.656 & 1.000 & 0.906 & 0.947 & 0.956 & 0.983 & 0.978 & 0.996 & 0.970 \\
    ferrite\_bead                & 0.529 & 0.914 & 0.408 & 0.806 & 0.634 & 0.717 & 0.894 & 0.817 & 0.913 & 0.932 & 0.965 & 0.966 & 0.967 & 0.978 \\
    fork\_crimp\_terminal        & 0.462 & 0.657 & 0.416 & 0.945 & 0.422 & 0.62  & 0.595 & 0.773 & 0.769 & 0.952 & 0.780 & 0.964 & 0.819 & 0.946 \\
    fuse\_holder                 & 0.853 & 0.861 & 0.564 & 0.957 & 0.309 & 0.605 & 0.597 & 0.754 & 0.736 & 0.927 & 0.770 & 0.948 & 0.866  & 0.915 \\
    headphone\_jack\_socket      & 0.996 & 0.914 & 0.933 & 0.879 & 0.607 & 0.633 & 0.660 & 0.696 & 0.919 & 0.942 & 0.982 & 0.982 & 0.994 & 0.987 \\
    humidity\_sensor             & 0.781 & 0.836 & 0.737 & 0.89  & 0.644 & 0.562 & 0.565 & 0.723 & 0.689 & 0.933 & 0.717 & 0.958 & 0.78 & 0.969 \\
    knob\_cap                    & 0.637 & 0.893 & 0.672 & 0.879 & 0.656 & 0.425 & 0.919 & 0.656 & 0.903 & 0.958 & 0.925 & 0.938 & 0.931 & 0.947 \\
    lattice\_block\_plug         & 0.833 & 0.852 & 0.79  & 0.898 & 0.769 & 0.776 & 0.842 & 0.919 & 0.911 & 0.923 & 0.917 & 0.958 & 0.939 & 0.941 \\
    lego\_pin\_connector\_plate  & 0.828 & 0.877 & 0.857 & 0.947 & 0.361 & 0.482 & 0.847 & 0.629 & 0.662 & 0.759 & 0.681 & 0.734 & 0.891 & 0.889 \\
    lego\_propeller              & 0.615 & 0.739 & 0.939 & 0.799 & 0.348 & 0.62  & 0.471 & 0.703 & 0.540 & 0.727 & 0.530 & 0.773 & 0.739  & 0.863 \\
    limit\_switch                & 0.846 & 0.95  & 0.823 & 0.79  & 0.763 & 0.545 & 0.804 & 0.641 & 0.822 & 0.938 & 0.863 & 0.966 & 0.925 & 0.984 \\
    miniature\_lifting\_motor    & 0.402 & 0.799 & 0.402 & 0.76  & 0.717 & 0.435 & 0.766 & 0.467 & 0.948 & 0.962 & 0.975 & 0.991 & 0.823  & 0.961 \\
    power\_jack                  & 0.354 & 0.664 & 0.176 & 0.489 & 0.433 & 0.687 & 0.564 & 0.645 & 0.981 & 0.923 & 0.996 & 0.902 & 0.973 & 0.947 \\
    purple\_clay\_pot            & 0.343 & 0.571 & 0.343 & 0.938 & 0.869 & 0.271 & 0.635 & 0.445 & 0.921 & 0.961 & 0.944 & 0.953 & 0.962 & 0.922 \\
    telephone\_spring\_switch    & 0.575 & 0.91  & 0.627 & 0.916 & 0.771 & 0.413 & 0.951 & 0.551 & 0.827 & 0.944 & 0.856 & 0.936 & 0.934 & 0.957 \\
    Avg                          & 0.645 & 0.808 & 0.659 & 0.843 & 0.644 & 0.554 & 0.693 & 0.650 & 0.812 & 0.905 & 0.841 & 0.922 & \textbf{0.890} & \textbf{0.937} \\
    \bottomrule
    \end{tabular}
  \label{tab2}}%
\end{table*}%

\begin{table*}[h]
  \centering
  \caption{Comparison of anomaly detection performance (I-AUROC and P-AUROC) across different modal combinations using PatchCore, M3DM, and our proposed method, evaluating the impact of 2D, Pesudo-3D, and 3D data integration.}
  \resizebox{0.7\textwidth}{!}{
    \begin{tabular}{cccccccccc}
    \toprule
    \multicolumn{2}{c}{PatchCore(2D)~\cite{patchcore}} & \multicolumn{2}{c}{PatchCore(Pesudo-3D)~\cite{patchcore}} & \multicolumn{2}{c}{M3DM(2D+3D)~\cite{wang2023multimodal}} & \multicolumn{2}{c}{PatchCore(2D+Pesudo-3D)~\cite{patchcore}} & \multicolumn{2}{c}{Ours(2D+Pesudo-3D+3D)} \\
    \midrule
    I-AUROC & P-AUROC & I-AUROC & P-AUROC & I-AUROC & P-AUROC & I-AUROC & P-AUROC & I-AUROC & P-AUROC \\
    \midrule
    0.747 & 0.875 & 0.715 & 0.681 & 0.752& 0.687& 0.839 & 0.867& \textbf{0.890}& \textbf{0.937}\\
    \bottomrule
    \end{tabular}%
  \label{tab3}}%
\end{table*}%
Single-modality methods show limitations: 2D struggles with depth-related defects like scratches and dents, while 3D lacks surface texture details. The 2D+3D combination provides complementary information, yet still misses subtle surface irregularities.
 Our D³M approach captures surface orientation and depth with pseudo-3D photometric stereo data, enhancing the detection of texture-based anomalies.

Figure~\ref{vis1} presents a comprehensive visual comparison across different modalities, demonstrating the significantly improved pixel-level anomaly detection performance of our method over the M3DM baseline, which utilizes only 2D and 3D data. By incorporating pseudo-3D information, our approach captures richer surface textures and enhanced depth details, enabling more precise and accurate identification of subtle defects that may otherwise be missed. This enhancement underscores the value of integrating pseudo-3D features in multimodal frameworks, providing a more robust solution for detecting complex and fine-grained anomalies in industrial applications.

\begin{figure}[h]
  \centering
  \includegraphics[width=0.48\textwidth]{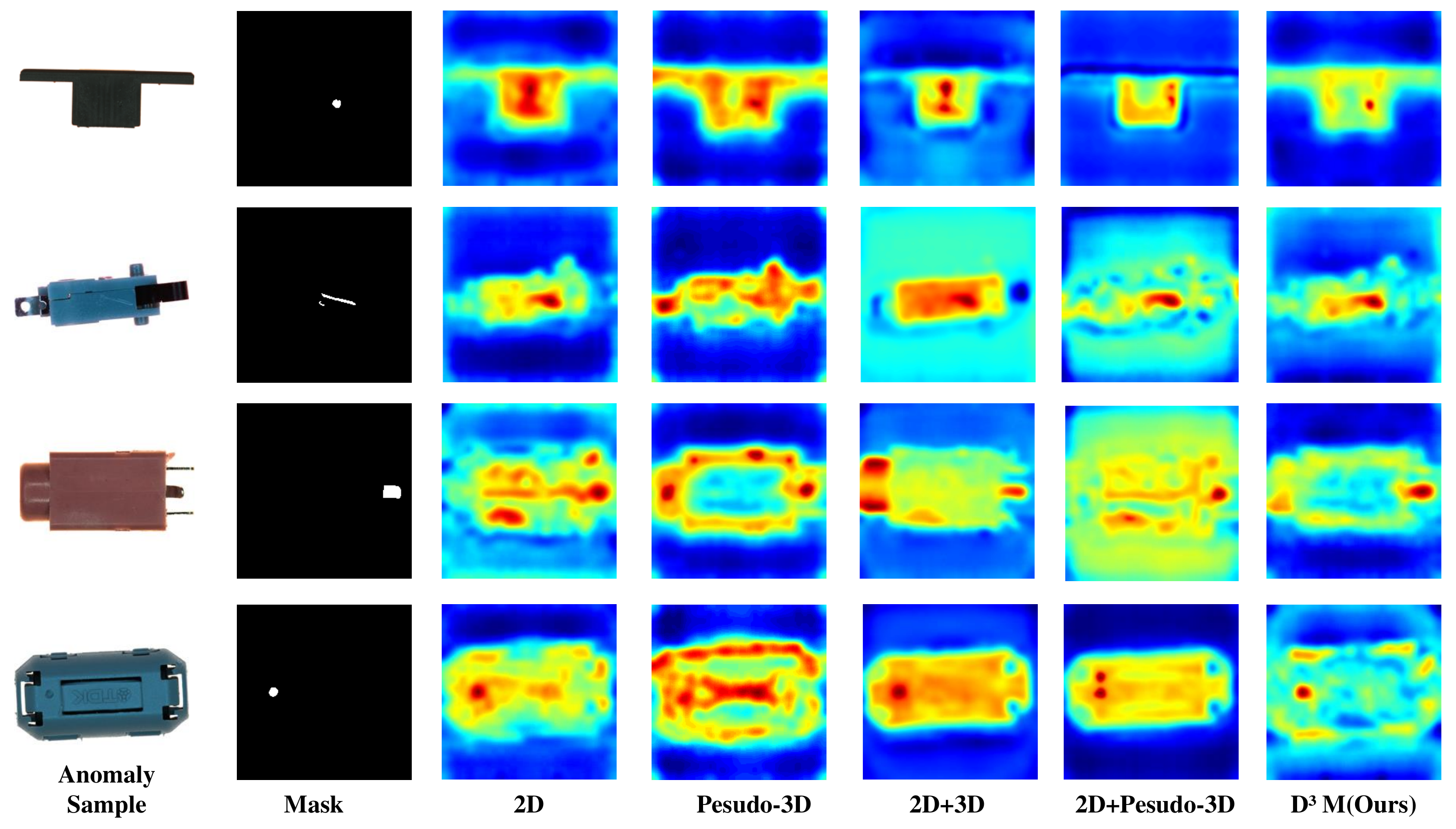}
  \caption{Visualization of segmentation map across different modality combinations.}
\label{vis2}
\vspace{-1em}
\end{figure}
\subsection{Analysis of Modality Combinations} 

Table~\ref{tab3} presents a comparative analysis of anomaly detection performance (I-AUROC and P-AUROC) across different modality combinations using PatchCore, M3DM, and D³M. By integrating 2D, pseudo-3D, and 3D data, our approach achieves the highest performance. Notably, the 2D+pseudo-3D combination already shows strong results by capturing essential surface details, while the addition of 3D data further enhances depth information, boosting detection accuracy. These results underscore the importance of pseudo-3D in multimodal fusion and the complementary role of 3D data for robust industrial anomaly detection.
In Figure~\ref{vis2}, we present the visual results across various modality combinations, which further validate the experimental findings. The Pseudo-3D modality highlights surface irregularities effectively, especially subtle textures and minor surface defects, while the addition of 2D enhances the clarity of these details. The 2D+pseudo-3D combination captures essential surface features, achieving strong results as shown in the table. Incorporating 3D data adds valuable depth information, allowing our D³M model to reach the highest accuracy.


\subsection{Effect of Point Cloud Resolution}
\begin{figure}[h]
  \centering
  \includegraphics[width=0.47\textwidth]{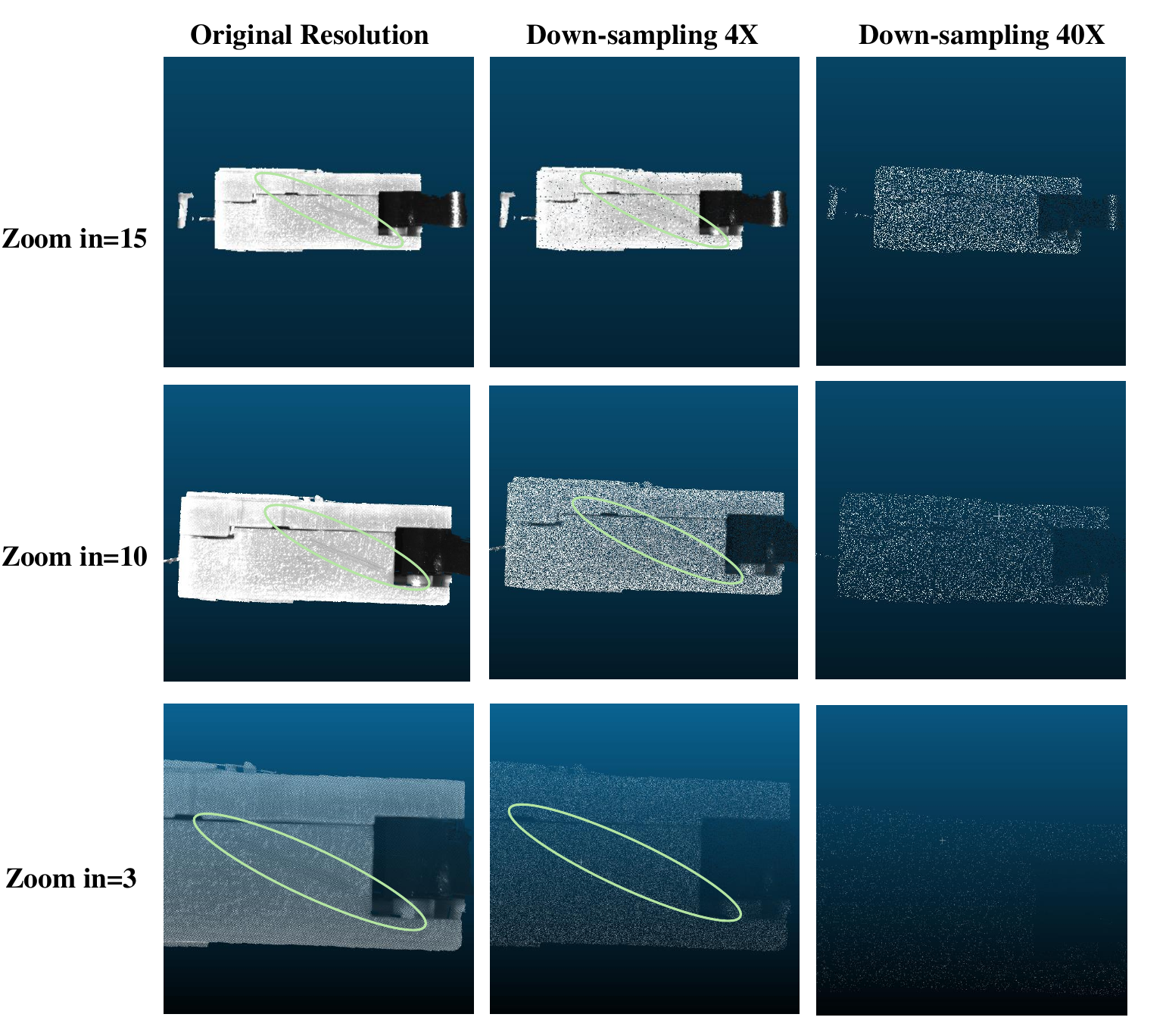}
  \caption{Visualization of point cloud data at original, 4x, and 40x downsampling, highlighting the impact of resolution reduction on detecting fine defects in smaller industrial components.}
\label{down}
  \vspace{-1em}
\end{figure}

As shown in Fig.~\ref{down}, we conducted downsampling experiments on point cloud data, reducing the original resolution by factors of 4x and 40x to simulate Real3D-AD and MVTec 3D-AD resolutions. Since our components are about one-tenth the size of those in MVTec 3D-AD and Real3D-AD, lower resolutions significantly hinder defect detection. Observing the performance at different scales (Raw, 4x, and 40x), subtle flaws visible at full resolution become hard to detect or vanish entirely at 40x downsampling, emphasizing the need for high-resolution data to capture fine details in industrial scenarios with smaller parts.

\begin{table}[htbp]
  \centering
  \caption{Performance of PointMAE for different point cloud resolutions (Original vs. 4x).}
      \resizebox{0.45\textwidth}{!}{

    \begin{tabular}{cccccc}
    \toprule
    \multirow{2}[4]{*}{Object} & \multirow{2}[4]{*}{Defect} & \multicolumn{2}{c}{I-AUROC} & \multicolumn{2}{c}{P-AUROC} \\
\cmidrule{3-6}          &       & Original   & 4x    & Original   & 4x \\
    \midrule
    \multirow{3}[2]{*}{connector\_housing-female} & Hole  & 0.516 & 0.500   & 0.819 & 0.801 \\
          & Deformation & 0.872 & 0.804 & 0.653 & 0.638 \\
          & Pit   & 1.000     & 0.976 & 0.457 & 0.447 \\
    \midrule
    \multirow{2}[2]{*}{telephone\_spring\_switch} & Damage & 0.724 & 0.648 & 0.396 & 0.376 \\
          & Hole  & 0.572 & 0.504 & 0.714 & 0.712 \\
    \midrule
    \multirow{2}[2]{*}{audio\_jack\_socket} & Impact Damage & 0.948 & 0.908 & 0.719 & 0.678 \\
          & Pit   & 0.972 & 0.948 & 0.893 & 0.874 \\
    \midrule
    \multirow{2}[2]{*}{headphone\_jack\_socket} & Scratch & 1.000     & 1.000     & 0.919 & 0.802 \\
          & Pit   & 1.000     & 1.000     & 0.905 & 0.832 \\
    \midrule
    limit-switch & Impact Damage & 1.000     & 0.922 & 0.843 & 0.821 \\
    \midrule
    fork\_crimp\_terminal & Scratch & 0.636 & 0.602 & 0.597 & 0.557 \\
    \bottomrule
    \end{tabular}%
  \label{tab4}}%
\end{table}%

The results in Table~\ref{tab4} show that downsampling the point cloud data leads to a noticeable performance decline, particularly for defects with small areas, such as ``Scratch'' and ``Pit''. For these defects, which have less than 0.1\% of the surface area affected, the resolution decrease significantly impairs the detection ability. 

\begin{table}[htbp]
  \centering
  \caption{Effect of feature interpolation on anomaly detection performance in the Real-IAD D³ and MVTec 3D-AD.}
  \resizebox{0.48\textwidth}{!}{
    \begin{tabular}{clll}
    \toprule
    \multirow{2}[4]{*}{Dataset} & \multicolumn{1}{c}{\multirow{2}[4]{*}{Feature interpolation}} & \multicolumn{2}{c}{M3DM~\cite{wang2023multimodal}} \\
\cmidrule{3-4}          &       & I-AUROC & P-AUROC \\
    \midrule
    \multirow{2}[2]{*}{MVTec 3D-AD} & w/ feature interpolation & 0.822 & 0.967 \\
          & w/o feature interpolation & 0.803(↓0.019) & 0.856(↓0.111) \\
    \midrule
    \multirow{2}[2]{*}{Real-IAD D³} & w/ feature interpolation & 0.841& 0.922\\
          & w/o feature interpolation & 0.832(↓0.009)& 0.934(↑0.012)\\
    \bottomrule
    \end{tabular}%
  \label{tab5}}%
    \vspace{-1em}

\end{table}%
Table~\ref{tab5} presents the effect of the interpolation module on anomaly detection performance in the MVTec 3D-AD and Real-IAD D3 data sets using the M3DM (DINO+PointMAE)~\cite{wang2023multimodal}. The M3DM interpolation module improves the density and representation of 3D point-cloud features through interpolation after backbone extraction. Experimental results show that in the MVTec 3D-AD dataset, removing the interpolation module leads to a notable drop in pixel-AUROC (by 0.111), indicating its contribution to maintaining feature density and extraction quality. However, in the Real-IAD D³ dataset, which inherently has higher-resolution point clouds, pixel-AUROC remains stable even without interpolation.
This confirms that high-resolution point clouds retain their original feature integrity post-backbone extraction, negating the need for interpolation. 

%% file: sec/5_conclusion.tex
\section{Conclusion}
In conclusion, this work introduces Real-IAD D³, a high-precision, large-scale multimodal dataset that integrates RGB, 3D point cloud, and pseudo-3D data to address limitations in existing industrial anomaly detection (IAD) datasets. 
Covering 20 product categories with diverse defects, Real-IAD D³ offers a realistic representation of industrial scenarios with enhanced scale, resolution, and diversity. The inclusion of pseudo-3D imagery improves surface and depth detail, significantly enhancing detection accuracy. 
Experimental results indicate that current methods face challenges in complex industrial settings, highlighting the need for more robust solutions.
Real-IAD D³ and its baseline establish a strong foundation for advancing multimodal IAD in industrial applications.


\section*{Acknowledgements}  
This work was partially supported by the National Natural Science Foundation of China (Grant Nos. 62171139 and 62302296), as well as the Shanghai Science and Technology Project (Grant No. 24YF2716900).


%% file: sec/X_suppl.tex
\clearpage
\renewcommand\thefigure{A\arabic{figure}}
\renewcommand\thetable{A\arabic{table}}  
\renewcommand\theequation{A\arabic{equation}}
\setcounter{equation}{0}
\setcounter{table}{0}
\setcounter{figure}{0}
\appendix

\section{Expanded Details of the Real-IAD D³ Dataset}

Table~\ref{tabsize} presents the dimensions of the materials included in the Real-IAD D³ dataset. In comparison with existing datasets, such as MVTec 3D-AD and Real3D-AD, the components in Real-IAD D³ are characterized by significantly smaller dimensions, which introduces unique challenges for anomaly detection tasks. Specifically, the materials in this dataset have lengths ranging from 7 mm to 27 mm, widths from 5 mm to 25 mm, and heights predominantly below 15 mm. These compact dimensions pose additional challenges for detecting subtle defects, as the anomalies often occupy only a small fraction of the material's surface, typically less than 3\% and in some cases as small as 0.46\%.

Furthermore, the materials in the dataset are sourced from real-world industrial components, including electronic devices, mechanical parts, and connectors. Examples include humidity sensors, audio jack sockets, fork crimp terminals, and ethernet connectors. This diversity in material types and geometries ensures the practical relevance of the dataset for industrial applications, reflecting real-world conditions where anomalies can vary significantly in appearance and location.

The combination of small material sizes and fine-grained defects, such as scratches, dents, and pits, considerably amplifies the difficulty of the anomaly detection task. These defects, which are often barely perceptible, demand high-resolution imaging and precise algorithms to capture the subtle variations in surface texture and geometry. The Real-IAD D³ dataset thus provides a rigorous benchmark for advancing multimodal anomaly detection in complex industrial settings.

\begin{table}[htbp]
  \centering
    \caption{Visualization of additional defects and corresponding products across 2D, pseudo-3D, and 3D modalities, showcasing the complementary strengths of each modality in capturing diverse defect characteristics.}
    \resizebox{0.48\textwidth}{!}{
    \begin{tabular}{cccc}
    \toprule
    Material Name & Length (mm) & Width (mm) & Height (mm) \\
    \midrule
    humidity\_sensor & 23    & 8     & 3 \\
    fuse\_holder & 27    & 10    & 7 \\
    ferrite\_bead & 23    & 10    & 10 \\
    lego\_pin\_connector\_plate & 15    & 8     & 3 \\
    fork\_crimp\_terminal & 22    & 5     & 5 \\
    purple\_clay\_pot & 20    & 20    & 8 \\
    ethernet\_connector & 17    & 13    & 10 \\
    miniature\_lifting\_motor & 23    & 20    & 4 \\
    dc\_power\_connector & 25    & 22    & 7 \\
    lego\_propeller & 25    & 25    & 10 \\
    limit\_switch & 17    & 8     & 6 \\
    headphone\_jack\_socket & 18    & 9     & 5 \\
    audio\_jack\_socket & 15    & 12    & 15 \\
    connector\_housing-female & 15    & 12    & 5 \\
    common-mode-filter & 10    & 10    & 12 \\
    lattice\_block\_plug & 16    & 12    & 15 \\
    knob\_cap & 7     & 7     & 5 \\
    telephone\_spring\_switch & 23    & 14    & 10 \\
    power\_jack & 15    & 12    & 17 \\
    crimp\_st\_cable\_mount\_box & 15    & 10    & 15 \\
    \bottomrule
    \end{tabular}%
  \label{tabsize}}%
\end{table}%

\section{Analysis of Additional Defects and Modalities in Real-IAD D³ Dataset}

\begin{figure*}[h]
    \centering
    \includegraphics[width=0.99\textwidth]{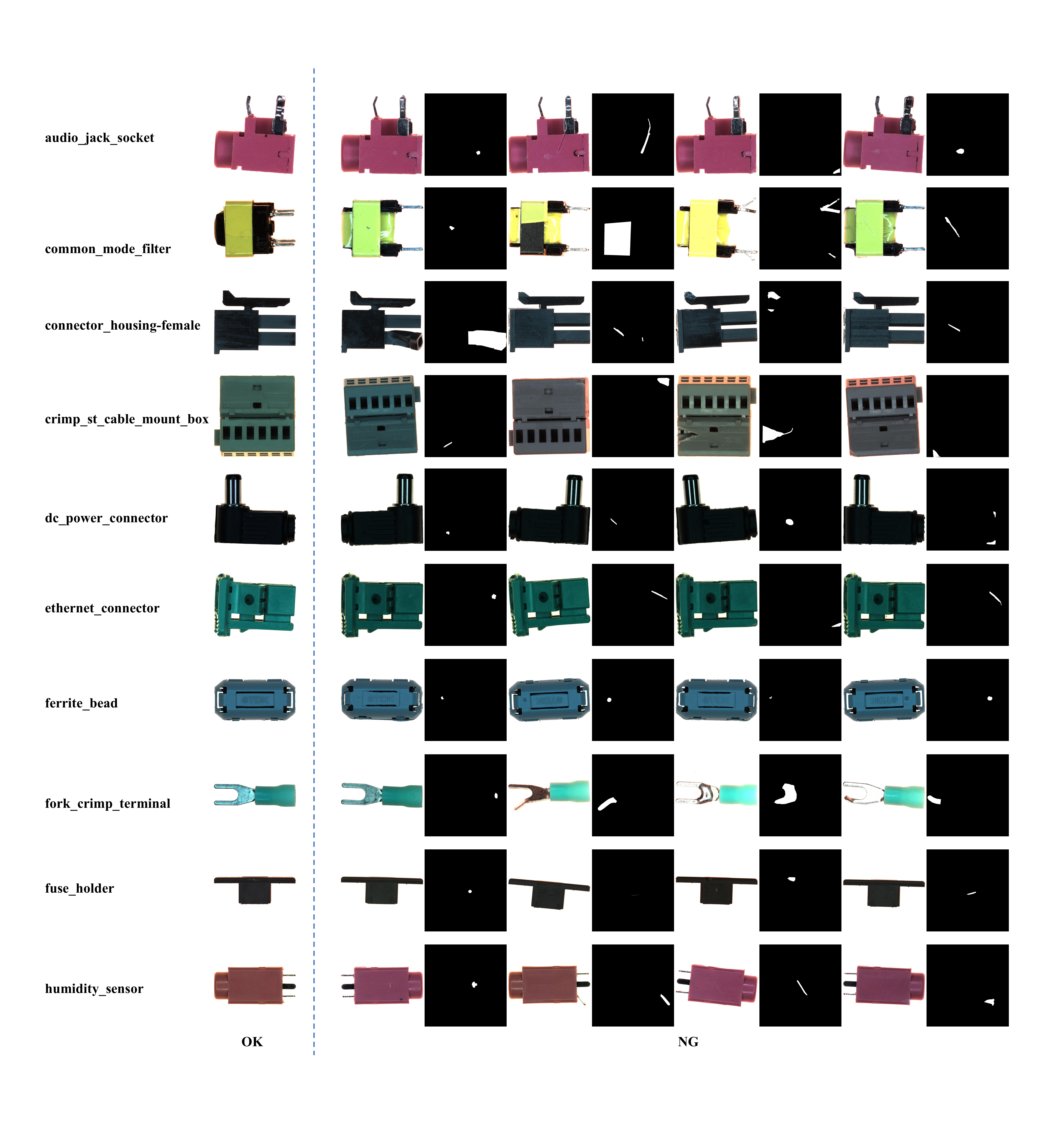}
\caption{Visualization of additional defects and their corresponding masks for the first ten product categories in the Real-IAD D³ dataset, showcasing the dataset's diversity and precision.}
    \label{sup_fig2}
\end{figure*}

Figure~\ref{sup_fig2} provides examples of defects and their corresponding masks for the first ten product categories. These examples demonstrate the diversity of materials and the high accuracy of defect annotations in the dataset. The displayed components, ranging from electronic connectors to mechanical parts, contain various types of surface anomalies such as scratches, dents, and cracks. The provided masks precisely delineate the defective regions, which are essential for both supervised training and objective evaluation of anomaly detection models.

\begin{figure*}[h]
    \centering
    \includegraphics[width=0.99\textwidth]{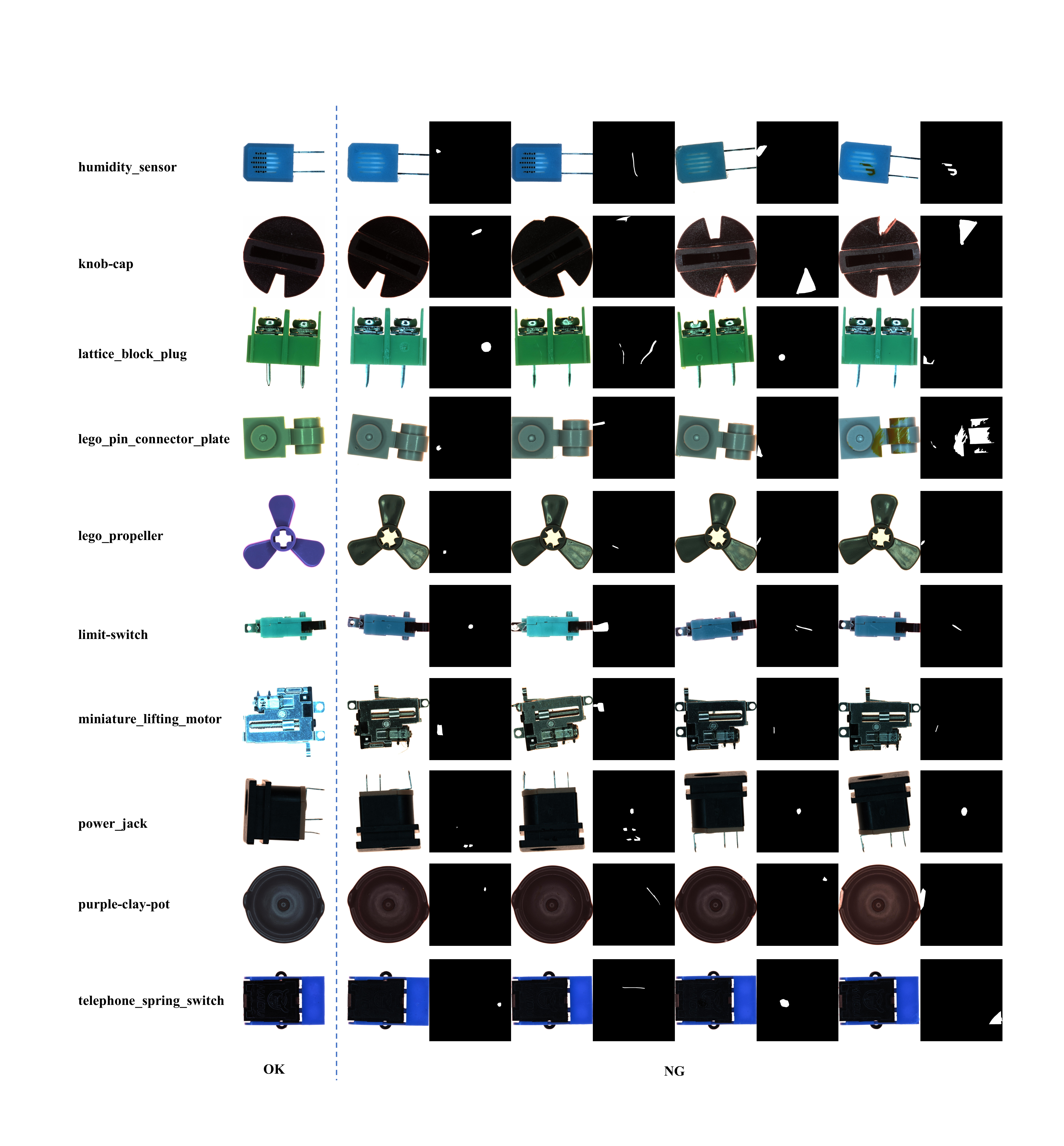}
\caption{Visualization of additional defects and corresponding masks for the second group of ten product categories in the Real-IAD D³ dataset, further illustrating its diversity and precision.}
    \label{sup_fig3}
\end{figure*}

Figure~\ref{sup_fig3} complements the previous set by presenting additional examples of defects and masks from another ten product categories. These categories feature a broader variety of geometries and textures, making the detection task more complex. The annotations continue to exhibit a high level of precision, supporting robust training and reliable benchmarking of detection algorithms. The combination of detailed annotations and diverse materials makes this dataset an excellent benchmark for evaluating anomaly detection methods in realistic industrial scenarios.

\begin{figure*}[h]
    \centering
    \includegraphics[width=0.99\textwidth]{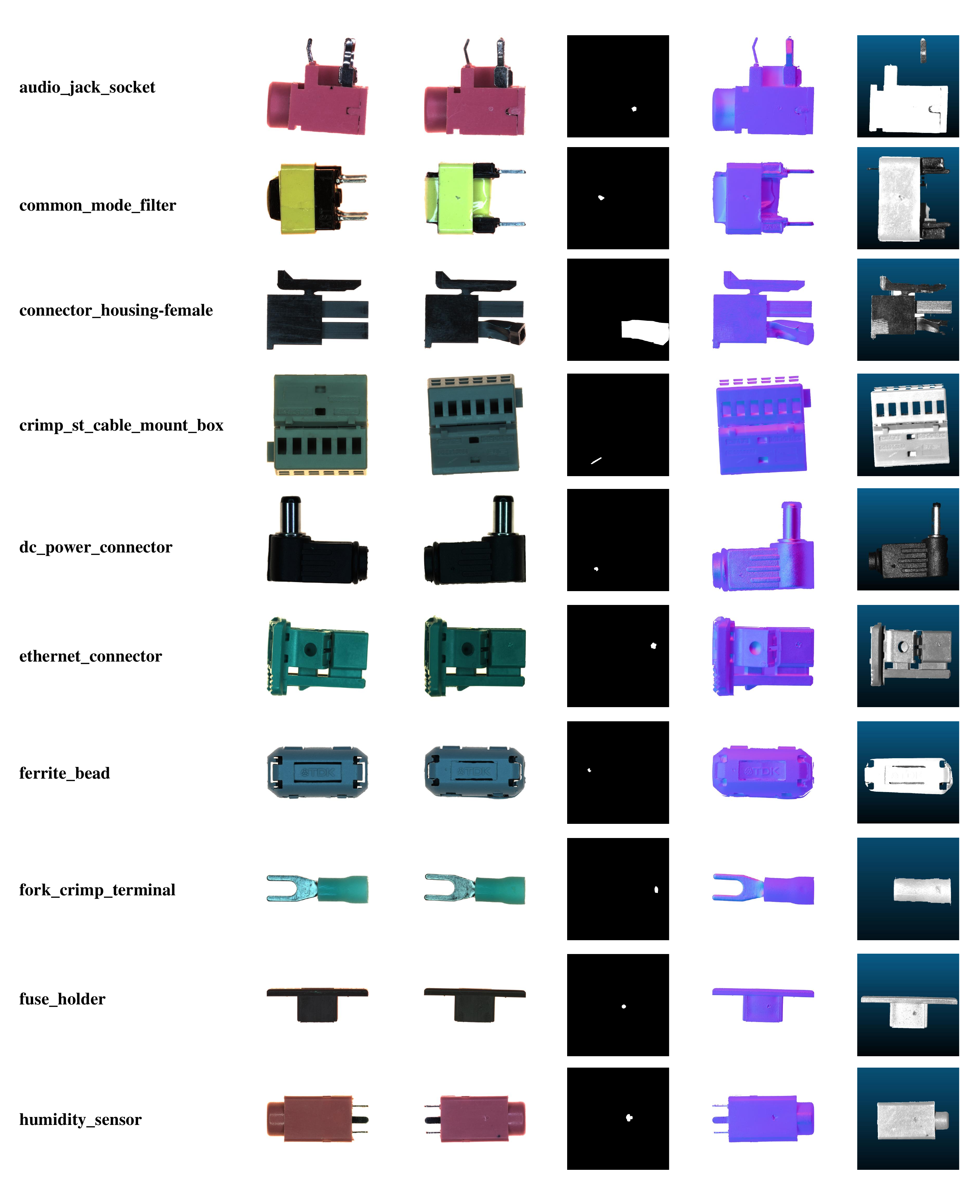}
    \caption{Visualization of additional defects and corresponding products across 2D, pseudo-3D, and 3D modalities, showcasing the complementary strengths of each modality in capturing diverse defect characteristics.}
    \label{sub_fig1}
\end{figure*}

Figure~\ref{sub_fig1} highlights the multimodal approach of the dataset, showing the integration of 2D images, pseudo-3D data, and 3D point clouds. The 2D images provide essential visual details such as surface texture and color variation, which are effective for identifying shallow defects like surface scratches. Pseudo-3D data captures depth variations, making it suitable for detecting surface irregularities such as dents that are difficult to perceive in standard 2D images. Finally, the 3D point clouds offer precise geometric information, which is invaluable for localizing structural defects such as cracks or deformations. Together, these modalities complement each other, providing a comprehensive framework for detecting a wide range of anomalies in industrial applications.

\section{Imaging Report Analysis}
\begin{figure*}[h]
    \centering
    \includegraphics[width=0.9\textwidth]{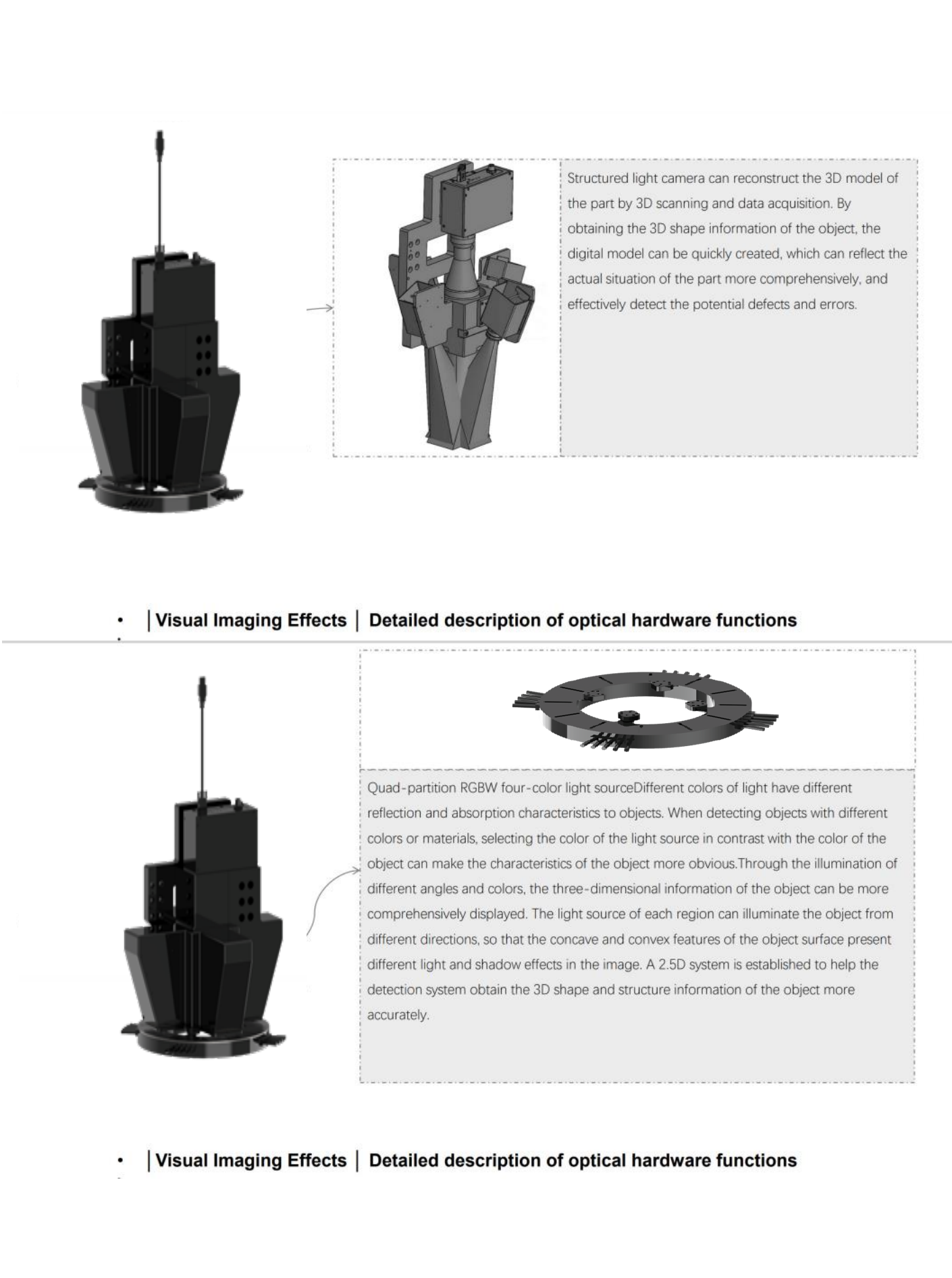}
\caption{Imaging report generated using the four-eye structured light system, demonstrating the captured raw data and its corresponding structured light patterns. The report showcases the effectiveness of the four-view system in capturing surface details and resolving occlusions.}
    \label{sub_fig4}
\end{figure*}

\begin{figure*}[h]
    \centering
    \includegraphics[width=0.9\textwidth]{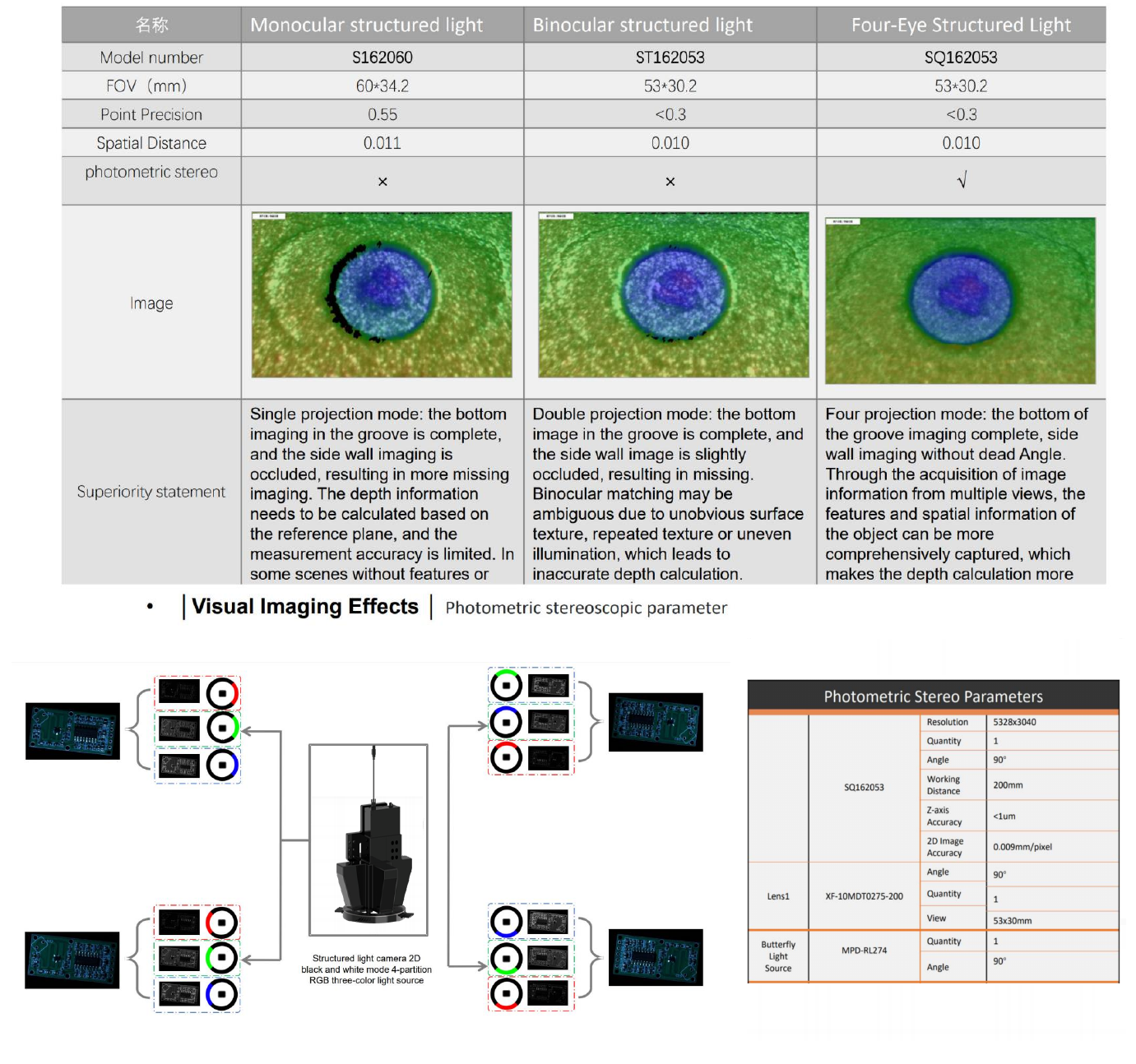}
\caption{Comparison of 3D imaging and pseudo-3D imaging results. The report highlights the differences in depth reconstruction and surface detail representation between the two modalities, illustrating the complementary strengths of pseudo-3D imaging for fine surface textures and 3D imaging for volumetric features.}
    \label{sub_fig5}
\end{figure*}

Figures~\ref{sub_fig4} and \ref{sub_fig5} present the imaging report generated from the experiments conducted using the proposed four-eye structured light system and its comparison with alternative imaging modalities. These reports comprehensively evaluate the system's capability in capturing surface details, resolving occlusions, and reconstructing accurate 3D models of industrial components.

%% file: main.bbl
\begin{thebibliography}{35}
\providecommand{\natexlab}[1]{#1}
\providecommand{\url}[1]{\texttt{#1}}
\expandafter\ifx\csname urlstyle\endcsname\relax
  \providecommand{\doi}[1]{doi: #1}\else
  \providecommand{\doi}{doi: \begingroup \urlstyle{rm}\Url}\fi

\bibitem[Bergmann and Sattlegger(2023)]{bergmann2023anomaly}
Paul Bergmann and David Sattlegger.
\newblock Anomaly detection in 3d point clouds using deep geometric descriptors.
\newblock In \emph{Proceedings of the IEEE/CVF Winter Conference on Applications of Computer Vision}, pages 2613--2623, 2023.

\bibitem[Bergmann et~al.(2019)Bergmann, Fauser, Sattlegger, and Steger]{bergmann2019mvtec}
Paul Bergmann, Michael Fauser, David Sattlegger, and Carsten Steger.
\newblock Mvtec ad--a comprehensive real-world dataset for unsupervised anomaly detection.
\newblock In \emph{Proceedings of the IEEE/CVF conference on computer vision and pattern recognition}, pages 9592--9600, 2019.

\bibitem[Bergmann et~al.(2020)Bergmann, Fauser, Sattlegger, and Steger]{bergmann2020uninformed}
Paul Bergmann, Michael Fauser, David Sattlegger, and Carsten Steger.
\newblock Uninformed students: Student-teacher anomaly detection with discriminative latent embeddings.
\newblock In \emph{Proceedings of the IEEE/CVF conference on computer vision and pattern recognition}, pages 4183--4192, 2020.

\bibitem[Bergmann et~al.(2021)Bergmann, Jin, Sattlegger, and Steger]{bergmann2021mvtec}
Paul Bergmann, Xin Jin, David Sattlegger, and Carsten Steger.
\newblock The mvtec 3d-ad dataset for unsupervised 3d anomaly detection and localization.
\newblock \emph{arXiv preprint arXiv:2112.09045}, 2021.

\bibitem[Bonfiglioli et~al.(2022)Bonfiglioli, Toschi, Silvestri, Fioraio, and De~Gregorio]{eyecandies}
Luca Bonfiglioli, Marco Toschi, Davide Silvestri, Nicola Fioraio, and Daniele De~Gregorio.
\newblock The eyecandies dataset for unsupervised multimodal anomaly detection and localization.
\newblock In \emph{Proceedings of the Asian Conference on Computer Vision}, pages 3586--3602, 2022.

\bibitem[Cao et~al.(2024)Cao, Xu, Zhang, Cheng, Huang, Pang, and Shen]{cao2024survey}
Yunkang Cao, Xiaohao Xu, Jiangning Zhang, Yuqi Cheng, Xiaonan Huang, Guansong Pang, and Weiming Shen.
\newblock A survey on visual anomaly detection: Challenge, approach, and prospect.
\newblock \emph{arXiv preprint arXiv:2401.16402}, 2024.

\bibitem[Chen et~al.(2023{\natexlab{a}})Chen, Xie, Liu, Wang, Luo, Wang, and Zheng]{chen2023easynet}
Ruitao Chen, Guoyang Xie, Jiaqi Liu, Jinbao Wang, Ziqi Luo, Jinfan Wang, and Feng Zheng.
\newblock Easynet: An efficient network for 3d industrial anomaly detection.
\newblock In \emph{Proceedings of the ACM International Conference on Multimedia}, pages 7038--7046, 2023{\natexlab{a}}.

\bibitem[Chen et~al.(2023{\natexlab{b}})Chen, Han, and Zhang]{chen2023zero}
Xuhai Chen, Yue Han, and Jiangning Zhang.
\newblock A zero-/few-shot anomaly classification and segmentation method for cvpr 2023 vand workshop challenge tracks 1\&2: 1st place on zero-shot ad and 4th place on few-shot ad.
\newblock \emph{arXiv preprint arXiv:2305.17382}, 2023{\natexlab{b}}.

\bibitem[Chu et~al.(2023)Chu, Liu, Hsieh, Chen, and Liu]{chu2023shape}
Yu-Min Chu, Chieh Liu, Ting-I Hsieh, Hwann-Tzong Chen, and Tyng-Luh Liu.
\newblock Shape-guided dual-memory learning for 3d anomaly detection.
\newblock In \emph{Proceedings of the 40th International Conference on Machine Learning}, pages 6185--6194, 2023.

\bibitem[Defard et~al.(2021)Defard, Setkov, Loesch, and Audigier]{defard2021padim}
Thomas Defard, Aleksandr Setkov, Angelique Loesch, and Romaric Audigier.
\newblock Padim: a patch distribution modeling framework for anomaly detection and localization.
\newblock In \emph{International Conference on Pattern Recognition}, pages 475--489. Springer, 2021.

\bibitem[Deng and Li(2022)]{deng2022anomaly}
Hanqiu Deng and Xingyu Li.
\newblock Anomaly detection via reverse distillation from one-class embedding.
\newblock In \emph{Proceedings of the IEEE/CVF Conference on Computer Vision and Pattern Recognition}, pages 9737--9746, 2022.

\bibitem[Gudovskiy et~al.(2022)]{gudovskiy2022cflow}
Denis Gudovskiy et~al.
\newblock Cflow-ad: Real-time unsupervised anomaly detection with compact neural flow.
\newblock \emph{IEEE Transactions on Artificial Intelligence}, page 2022, 2022.

\bibitem[He et~al.(2024{\natexlab{a}})He, Zhang, Chen, Chen, Li, Chen, Wang, Wang, and Xie]{diad}
Haoyang He, Jiangning Zhang, Hongxu Chen, Xuhai Chen, Zhishan Li, Xu Chen, Yabiao Wang, Chengjie Wang, and Lei Xie.
\newblock A diffusion-based framework for multi-class anomaly detection.
\newblock In \emph{Proceedings of the AAAI Conference on Artificial Intelligence}, pages 8472--8480, 2024{\natexlab{a}}.

\bibitem[He et~al.(2024{\natexlab{b}})He, Jiang, Peng, Liu, Du, Hu, Zhu, Chi, Wang, and Wang]{rlr}
Liren He, Zhengkai Jiang, Jinlong Peng, Liang Liu, Qiangang Du, Xiaobin Hu, Wenbing Zhu, Mingmin Chi, Yabiao Wang, and Chengjie Wang.
\newblock Learning unified reference representation for unsupervised multi-class anomaly detection.
\newblock \emph{arXiv preprint arXiv:2403.11561}, 2024{\natexlab{b}}.

\bibitem[Horwitz and Hoshen(2023)]{horwitz2023back}
Eliahu Horwitz and Yedid Hoshen.
\newblock Back to the feature: classical 3d features are (almost) all you need for 3d anomaly detection.
\newblock In \emph{Proceedings of the IEEE/CVF Conference on Computer Vision and Pattern Recognition}, pages 2967--2976, 2023.

\bibitem[Jeong et~al.(2023)Jeong, Zou, Kim, Zhang, Ravichandran, and Dabeer]{winclip}
Jongheon Jeong, Yang Zou, Taewan Kim, Dongqing Zhang, Avinash Ravichandran, and Onkar Dabeer.
\newblock Winclip: Zero-/few-shot anomaly classification and segmentation.
\newblock In \emph{Proceedings of the IEEE/CVF Conference on Computer Vision and Pattern Recognition}, pages 19606--19616, 2023.

\bibitem[Lei et~al.(2023)Lei, Hu, Wang, and Liu]{lei2023pyramidflow}
Jiarui Lei, Xiaobo Hu, Yue Wang, and Dong Liu.
\newblock Pyramidflow: High-resolution defect contrastive localization using pyramid normalizing flow.
\newblock In \emph{Proceedings of the IEEE/CVF Conference on Computer Vision and Pattern Recognition}, pages 14143--14152, 2023.

\bibitem[Li et~al.(2024)Li, Zhang, Tan, Chen, Qu, Xie, and Ma]{PromptAD}
Xiaofan Li, Zhizhong Zhang, Xin Tan, Chengwei Chen, Yanyun Qu, Yuan Xie, and Lizhuang Ma.
\newblock Promptad: Learning prompts with only normal samples for few-shot anomaly detection.
\newblock In \emph{Proceedings of the IEEE/CVF Conference on Computer Vision and Pattern Recognition}, pages 16838--16848, 2024.

\bibitem[Liu et~al.(2024)Liu, Xie, Chen, Li, Wang, Liu, Wang, and Zheng]{liu2024real3d}
Jiaqi Liu, Guoyang Xie, Ruitao Chen, Xinpeng Li, Jinbao Wang, Yong Liu, Chengjie Wang, and Feng Zheng.
\newblock Real3d-ad: A dataset of point cloud anomaly detection.
\newblock \emph{Advances in Neural Information Processing Systems}, 36, 2024.

\bibitem[Rippel et~al.(2021)Rippel, Mertens, and Merhof]{rippel2021modeling}
Oliver Rippel, Patrick Mertens, and Dorit Merhof.
\newblock Modeling the distribution of normal data in pre-trained deep features for anomaly detection.
\newblock In \emph{2020 25th International Conference on Pattern Recognition (ICPR)}, pages 6726--6733. IEEE, 2021.

\bibitem[Roth et~al.(2022{\natexlab{a}})Roth, Pemula, Zepeda, Sch\"olkopf, Brox, and Gehler]{patchcore}
Karsten Roth, Latha Pemula, Joaquin Zepeda, Bernhard Sch\"olkopf, Thomas Brox, and Peter Gehler.
\newblock Towards total recall in industrial anomaly detection.
\newblock In \emph{Proceedings of the IEEE/CVF Conference on Computer Vision and Pattern Recognition (CVPR)}, pages 14318--14328, 2022{\natexlab{a}}.

\bibitem[Roth et~al.(2022{\natexlab{b}})Roth, Pemula, Zepeda, Sch{\"o}lkopf, Brox, and Gehler]{roth2022towards}
Karsten Roth, Latha Pemula, Joaquin Zepeda, Bernhard Sch{\"o}lkopf, Thomas Brox, and Peter Gehler.
\newblock Towards total recall in industrial anomaly detection.
\newblock In \emph{Proceedings of the IEEE/CVF Conference on Computer Vision and Pattern Recognition}, pages 14318--14328, 2022{\natexlab{b}}.

\bibitem[Rudolph et~al.(2023)Rudolph, Wehrbein, Rosenhahn, and Wandt]{rudolph2023asymmetric}
Marco Rudolph, Tom Wehrbein, Bodo Rosenhahn, and Bastian Wandt.
\newblock Asymmetric student-teacher networks for industrial anomaly detection.
\newblock In \emph{Proceedings of the IEEE/CVF Winter Conference on Applications of Computer Vision}, pages 2592--2602, 2023.

\bibitem[Salehi et~al.(2021)Salehi, Sadjadi, Baselizadeh, Rohban, and Rabiee]{salehi2021multiresolution}
Mohammadreza Salehi, Niousha Sadjadi, Soroosh Baselizadeh, Mohammad~H Rohban, and Hamid~R Rabiee.
\newblock Multiresolution knowledge distillation for anomaly detection.
\newblock In \emph{Proceedings of the IEEE/CVF conference on computer vision and pattern recognition}, pages 14902--14912, 2021.

\bibitem[Tien et~al.(2023)Tien, Nguyen, Tran, Huy, Duong, Nguyen, and Truong]{tien2023revisiting}
Tran~Dinh Tien, Anh~Tuan Nguyen, Nguyen~Hoang Tran, Ta~Duc Huy, Soan Duong, Chanh D~Tr Nguyen, and Steven~QH Truong.
\newblock Revisiting reverse distillation for anomaly detection.
\newblock In \emph{Proceedings of the IEEE/CVF Conference on Computer Vision and Pattern Recognition}, pages 24511--24520, 2023.

\bibitem[Wan et~al.(2022)Wan, Cao, Gao, Shen, and Li]{wan2022position}
Qian Wan, Yunkang Cao, Liang Gao, Weiming Shen, and Xinyu Li.
\newblock Position encoding enhanced feature mapping for image anomaly detection.
\newblock In \emph{2022 IEEE 18th International Conference on Automation Science and Engineering (CASE)}, pages 876--881. IEEE, 2022.

\bibitem[Wang et~al.(2024{\natexlab{a}})Wang, Xu, Gan, Li, Hu, Zhu, and Ma]{wang2024pspu}
Chengjie Wang, Chengming Xu, Zhenye Gan, Yuxi Li, Jianlong Hu, Wenbing Zhu, and Lizhuang Ma.
\newblock Pspu: Enhanced positive and unlabeled learning by leveraging pseudo supervision.
\newblock In \emph{2024 IEEE International Conference on Multimedia and Expo (ICME)}, pages 1--6. IEEE, 2024{\natexlab{a}}.

\bibitem[Wang et~al.(2024{\natexlab{b}})Wang, Zhu, Gao, Gan, Zhang, Gu, Qian, Chen, and Ma]{wang2024real}
Chengjie Wang, Wenbing Zhu, Bin-Bin Gao, Zhenye Gan, Jiangning Zhang, Zhihao Gu, Shuguang Qian, Mingang Chen, and Lizhuang Ma.
\newblock Real-iad: A real-world multi-view dataset for benchmarking versatile industrial anomaly detection.
\newblock In \emph{Proceedings of the IEEE/CVF Conference on Computer Vision and Pattern Recognition}, pages 22883--22892, 2024{\natexlab{b}}.

\bibitem[Wang et~al.(2025)Wang, Jiang, Gao, Gan, Liu, Zheng, and Ma]{wang2025softpatch+}
Chengjie Wang, Xi Jiang, Bin-Bin Gao, Zhenye Gan, Yong Liu, Feng Zheng, and Lizhuang Ma.
\newblock Softpatch+: Fully unsupervised anomaly classification and segmentation.
\newblock \emph{Pattern Recognition}, 161:\penalty0 111295, 2025.

\bibitem[Wang et~al.(2023)Wang, Peng, Zhang, Yi, Wang, and Wang]{wang2023multimodal}
Yue Wang, Jinlong Peng, Jiangning Zhang, Ran Yi, Yabiao Wang, and Chengjie Wang.
\newblock Multimodal industrial anomaly detection via hybrid fusion.
\newblock In \emph{Proceedings of the IEEE/CVF Conference on Computer Vision and Pattern Recognition}, pages 8032--8041, 2023.

\bibitem[Xie et~al.(2022)]{xie2023graphcore}
Guoyang Xie et~al.
\newblock Pushing the limits of fewshot anomaly detection in industry vision: Graphcore.
\newblock \emph{International Conference on Learning Representations (ICLR)}, 2022.

\bibitem[You et~al.(2022)You, Cui, Shen, Yang, Lu, Zheng, and Le]{uniad}
Zhiyuan You, Lei Cui, Yujun Shen, Kai Yang, Xin Lu, Yu Zheng, and Xinyi Le.
\newblock A unified model for multi-class anomaly detection.
\newblock In \emph{Advances in Neural Information Processing Systems}, pages 4571--4584. Curran Associates, Inc., 2022.

\bibitem[Zavrtanik et~al.(2024)Zavrtanik, Kristan, and Sko{\v{c}}aj]{zavrtanik2024cheating}
Vitjan Zavrtanik, Matej Kristan, and Danijel Sko{\v{c}}aj.
\newblock Cheating depth: Enhancing 3d surface anomaly detection via depth simulation.
\newblock In \emph{Proceedings of the IEEE/CVF Winter Conference on Applications of Computer Vision}, pages 2164--2172, 2024.

\bibitem[Zhang et~al.(2023)]{zhang2023augmentation}
Lingrui Zhang et~al.
\newblock What makes a good data augmentation for few-shot unsupervised image anomaly detection?
\newblock In \emph{Proceedings of the IEEE/CVF Conference on Computer Vision and Pattern Recognition}, pages 4344--4353, 2023.

\bibitem[Zou et~al.(2022)Zou, Qiu, and Shen]{zou2022spot}
Zongyi Zou, Qiang Qiu, and Weiming Shen.
\newblock Spot-the-difference: A novel benchmark for image anomaly detection in industrial inspection.
\newblock In \emph{Proceedings of the IEEE/CVF International Conference on Computer Vision}, pages 2608--2616, 2022.

\end{thebibliography}
